\documentclass[preprint,10pt]{elsarticle}
\biboptions{sort&compress} 

\usepackage{amssymb}
\usepackage{amsmath}
\usepackage{graphicx}
\usepackage{float}
\usepackage{adjustbox}
\usepackage{makecell}
\journal{ }

\usepackage{multirow}
\usepackage{url}
\usepackage{enumitem}
\usepackage{setspace}

\usepackage[T1]{fontenc}
\usepackage{lmodern}
\usepackage[final]{microtype}
\newcounter{expitem}[subsection]
\newcounter{expsubitem}[expitem]

\newcolumntype{P}[1]{>{\centering\arraybackslash}p{#1}}
\newcolumntype{Q}[1]{>{\raggedright\arraybackslash}p{#1}}
\newcolumntype{M}[1]{>{\raggedright\arraybackslash}m{#1}}
\newcolumntype{L}[1]{>{\raggedright\let\newline\\\arraybackslash\hspace{0pt}}m{#1}}
\newcolumntype{C}[1]{>{\centering\let\newline\\\arraybackslash\hspace{0pt}}m{#1}}
\newcolumntype{R}[1]{>{\raggedleft\let\newline\\\arraybackslash\hspace{0pt}}m{#1}}
\newcolumntype{Z}[1]{>{\raggedright\arraybackslash}m{#1}} 
\newcolumntype{Y}[1]{>{\centering\arraybackslash}m{#1}}   
\usepackage{color,soul} 
\definecolor{color:major}{rgb}{0.0, 0.0, 0.0}
\definecolor{color:minor}{rgb}{0.0, 0.0, 0.0}
\definecolor{color:green}{rgb}{0.0, 0.0, 0.0}
\begin{document}

\begin{frontmatter}

\title{Personalized Object Identification and Localization via In-Context Inference with Vision-Language Models}

\author[label1]{Kensuke Nakamura}
\ead{kensuke@image.cau.ac.kr}

\author[label1]{Byung-Woo Hong\corref{cor1}}
\ead{hong@cau.ac.kr}

\address[label1]{Artificial Intelligence Department, Chung-Ang University, Seoul, 06974, Korea}

\cortext[cor1]{Corresponding author}

\begin{abstract}
Personalized object localization (POL) localizes an object instance in a query image based on a few reference images with bounding-box annotations and a target object label. The pioneering method, IPLoc, solves this task through in-context inference with vision-language models (VLMs). However, it assumes that the query image always contains the target object. This assumption severely limits its applicability to real-world scenarios with many irrelevant images. To address this issue, we formulate a new task, personalized object identification and localization (POIL), by positioning POL within the broader few-shot object detection framework. POIL aims to localize the target object instance while rejecting query images that do not contain the reference object instance. We also present POIL datasets constructed from public sources. We further propose an in-context algorithm named IPLoc-ID for solving POIL with VLMs. IPLoc-ID first predicts a candidate bounding box and then determines whether it corresponds to the reference object instance. We introduce a self-posed query to connect these two steps within a single autoregressive generation framework. Through ablation studies and comprehensive experiments, we show that IPLoc-ID substantially suppresses false-positive detections on negative query images while maintaining localization performance comparable to IPLoc. Overall, IPLoc-ID effectively addresses the practical instance-level POIL task, which cannot be sufficiently solved by conventional object detection, few-shot object detection, or the localization-only IPLoc method.
\end{abstract}
%

\end{frontmatter}


%

\clearpage
%
%
%
%
%
%
%
%
\section{Introduction} \label{sec:introduction}
Object detection (OD) is a fundamental visual recognition task that aims to find objects in an image and estimate their locations as bounding boxes. 
Recent advances in open-vocabulary object detection and few-shot object detection (FSOD) have made it possible to detect objects specified not only by predefined categories but also by text labels or a small number of support examples~\cite{minderer2022simple,liu2024groundingdino,kohler2023few,xin2024few,wang2020frustratingly}. 
However, most of these methods are essentially designed for category-level detection and do not aim to identify a specific object instance indicated by reference data. 
For example, even when a reference image specifies a particular cat, conventional OD or FSOD methods may regard detecting another cat from the same category as a successful result. 
In contrast, reference-conditioned instance-level localization aims to detect a specific object instance indicated by reference data in a query image.
Such a capability is expected to be useful for future applications such as user-specified image retrieval, video grounding, object re-identification, and personalized object tracking.
\par
In this line of research, IPLoc (in-context personalized object localization)~\cite{doveh2025iploc} is pioneering work on reference-conditioned instance-level localization.
It exploits the contextual understanding ability of transformer-based vision-language models (VLMs) to localize the corresponding object region in a query image based on reference data.
IPLoc takes a small number of images with bounding-box (BBOX) annotations and the target label as reference data, and generates the BBOX coordinates for the query image through next-token prediction. 
This formulation enables reference-conditioned inference with VLMs without fine-tuning to the reference data.
However, IPLoc assumes that the target object is present in the query image, i.e., the query image is positive. 
Therefore, even when the target object is absent from a negative query image, IPLoc still generates a bounding box. 
As a result, in practical scenarios such as image retrieval and video grounding, where most candidate images may not contain the object of interest, users must either preselect positive query images before inference or manually remove false-positive detections after inference.
This severely limits the practical applicability of the IPLoc framework.
%
%
%
%
%
\def \fw {55pt}
\def \pw {44pt}
\def \qw {40pt}
\begin{figure}[t]
\vspace{-12pt} 
\centering
\scriptsize
\hspace*{-16pt}
\begin{tabular}{P{\qw} p{2pt}  P{\pw}P{\pw}P{\pw}P{\pw}P{\pw}}
\multicolumn{1}{c}{\quad Reference data} & & \multicolumn{1}{r}{Florence-2} & \multicolumn{1}{c}{\quad NT3} & \multicolumn{1}{r}{Qwen2-VL} & \quad IPLoc & \quad IPLoc-ID  \\
%
\multirow[c]{2}{*}
{
    \shortstack[c]{%
        \vspace{-18pt}\\
        \\
        \includegraphics[width=\fw]{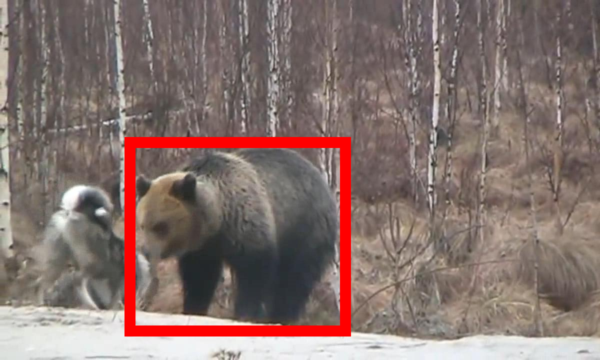}\\
        \scriptsize \textit{``bear''}
    }%
}
&
\rotatebox{90}{\parbox{10mm}{\centering  \vspace{3pt} \qquad \tiny Positive}} &
\includegraphics[width=\fw]{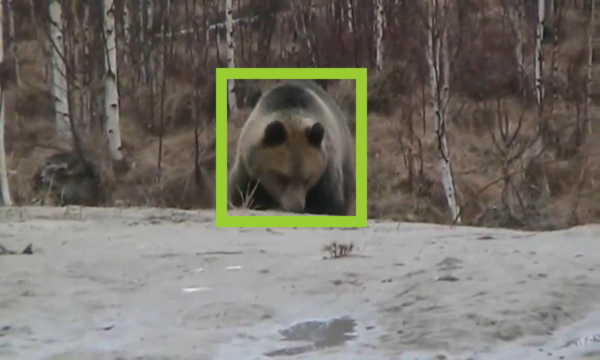} &
\includegraphics[width=\fw]{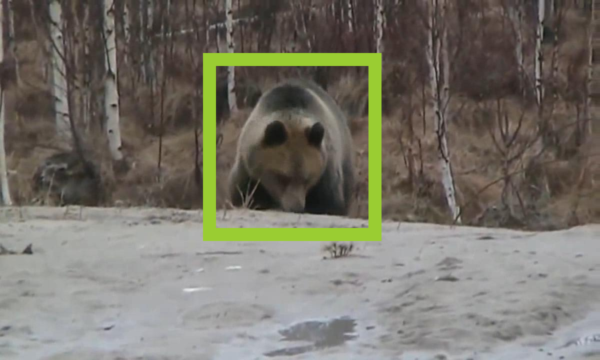} &
\includegraphics[width=\fw]{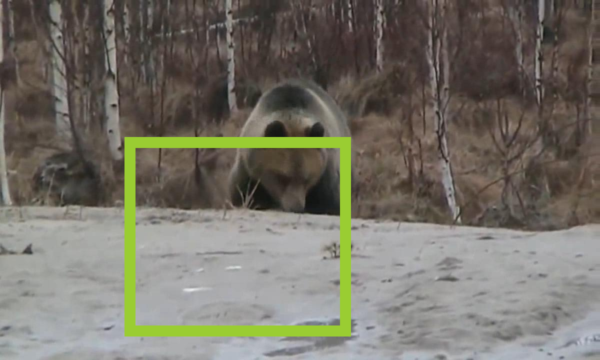} &
\includegraphics[width=\fw]{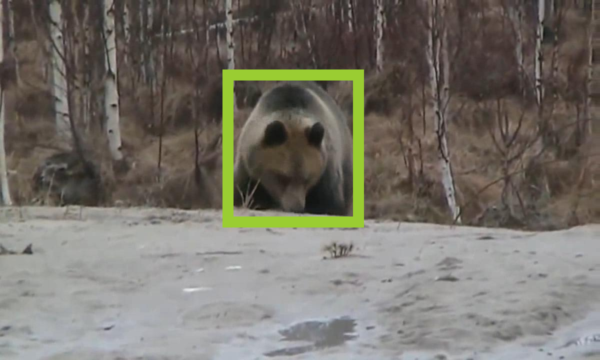} &
\includegraphics[width=\fw]{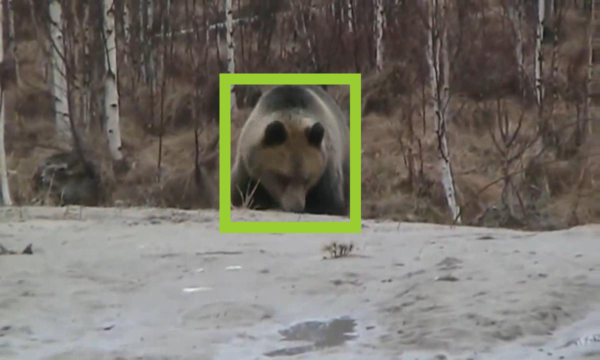} \\
&
\rotatebox{90}{\parbox{10mm}{\centering  \vspace{3pt} \quad \tiny Negative}} &
\includegraphics[width=\fw]{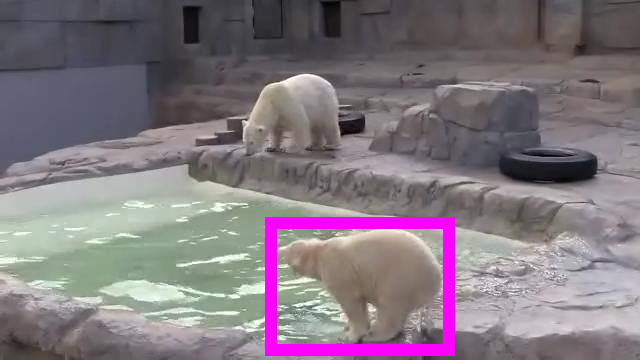} &
\includegraphics[width=\fw]{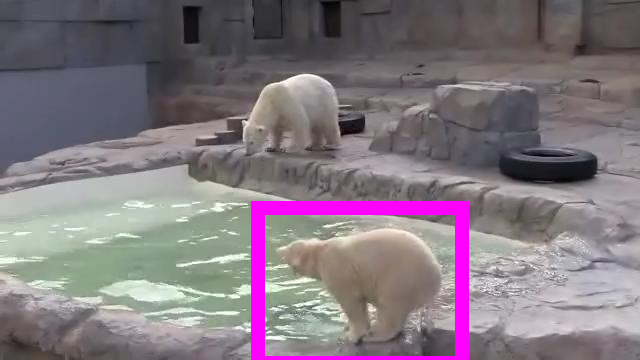} &
\includegraphics[width=\fw]{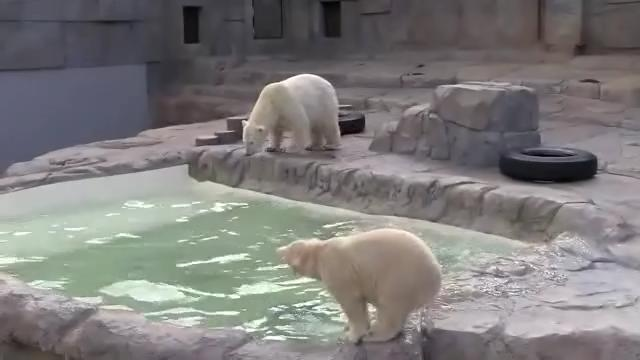} &
\includegraphics[width=\fw]{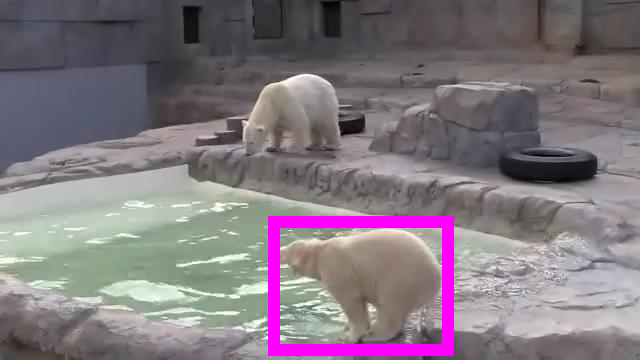} &
\includegraphics[width=\fw]{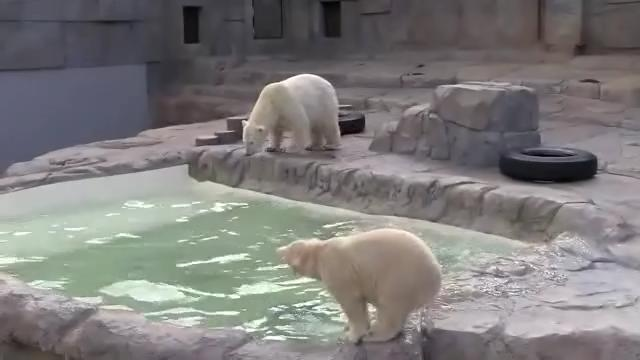} \\
\vspace{1pt}\\
%
\multirow[c]{2}{*}
    {
    \shortstack[c]{%
        \vspace{-18pt}\\
        \\
        \includegraphics[width=\fw]{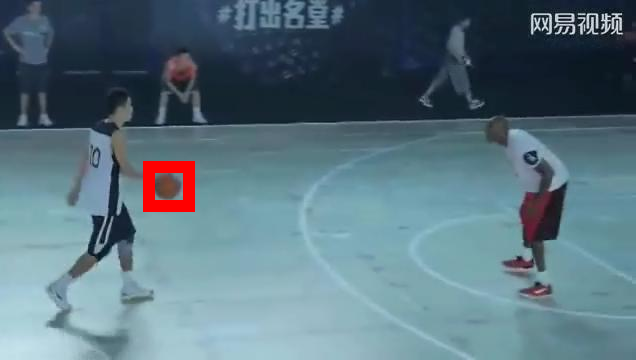}\\
        \scriptsize \textit{``basketball''}
    }%
}
&
\rotatebox{90}{\parbox{10mm}{\centering  \vspace{3pt} \quad \tiny Positive}} &
\includegraphics[width=\fw]{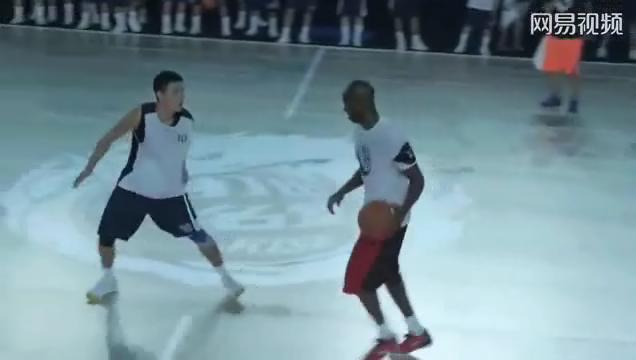} &
\includegraphics[width=\fw]{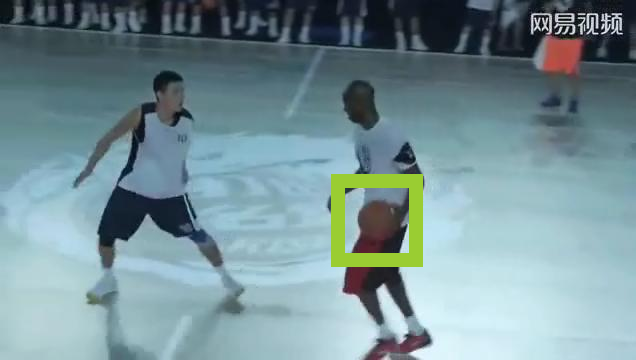} &
\includegraphics[width=\fw]{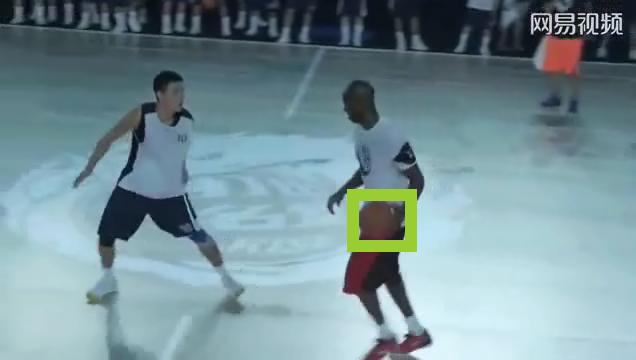} &
\includegraphics[width=\fw]{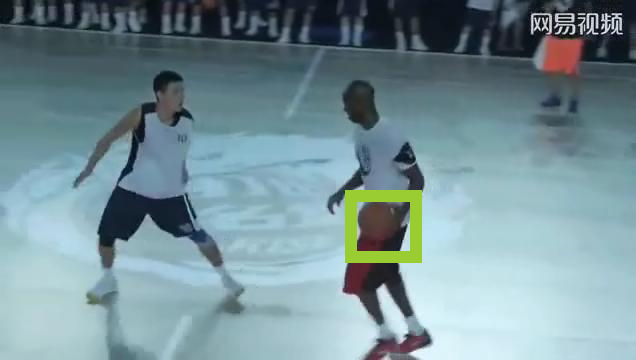} &
\includegraphics[width=\fw]{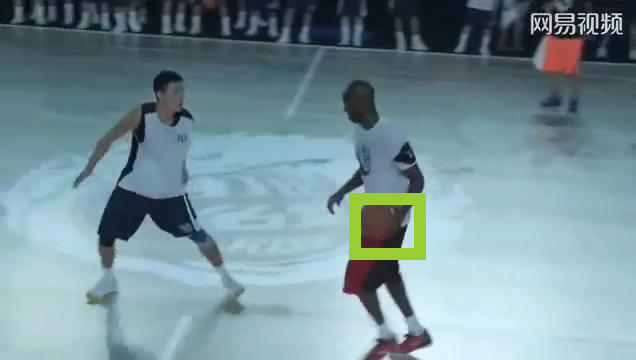} \\
 &
\rotatebox{90}{\parbox{10mm}{\centering  \vspace{3pt} \quad \tiny Negative}} &
\includegraphics[width=\fw]{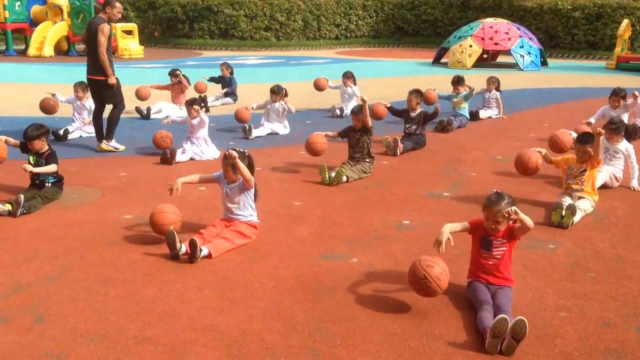} &
\includegraphics[width=\fw]{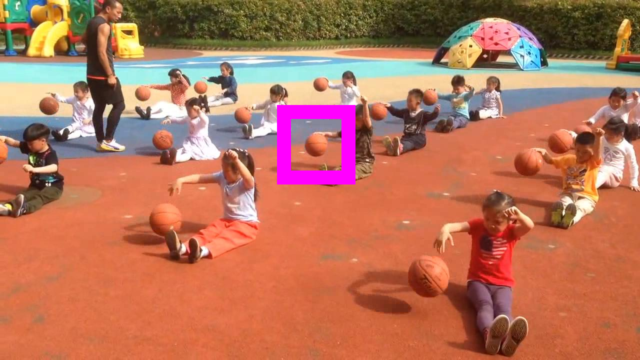} &
\includegraphics[width=\fw]{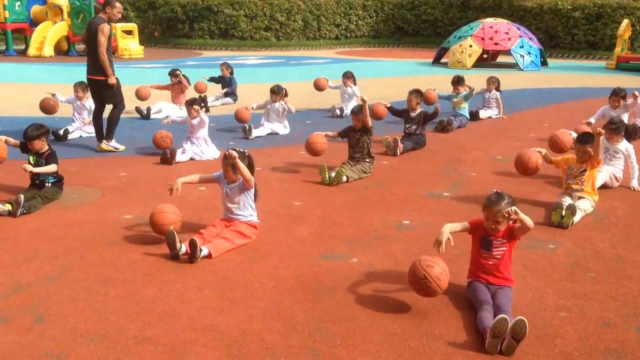} &
\includegraphics[width=\fw]{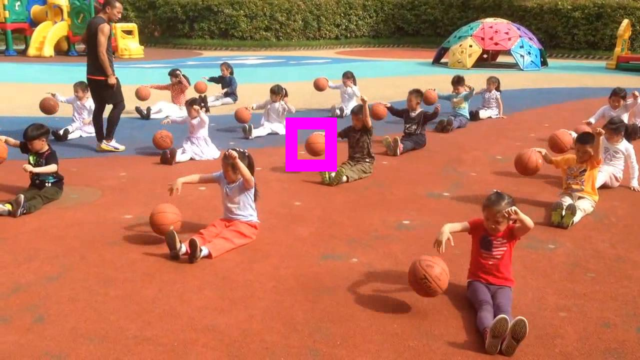} &
\includegraphics[width=\fw]{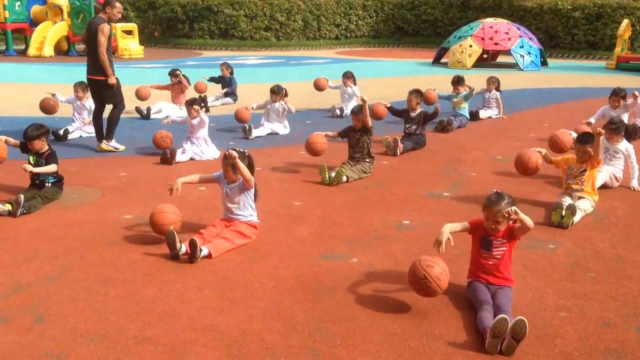} \\
\vspace{1pt}\\
%
\multirow[c]{2}{*}
{
\shortstack[c]{%
    \vspace{-18pt}\\
    \\
    \includegraphics[width=\fw]{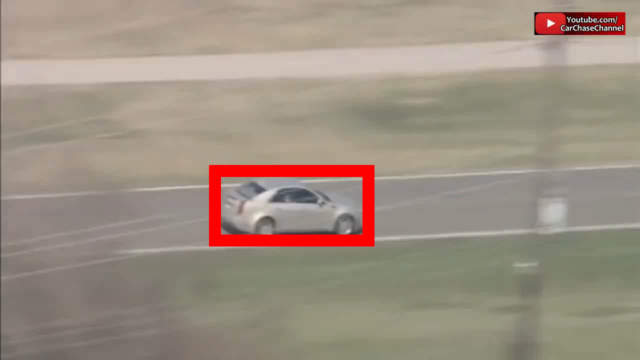}\\
    \scriptsize \textit{``car''}
}%
}
&
\rotatebox{90}{\parbox{10mm}{\centering  \vspace{3pt} \quad \tiny Positive}} &
\includegraphics[width=\fw]{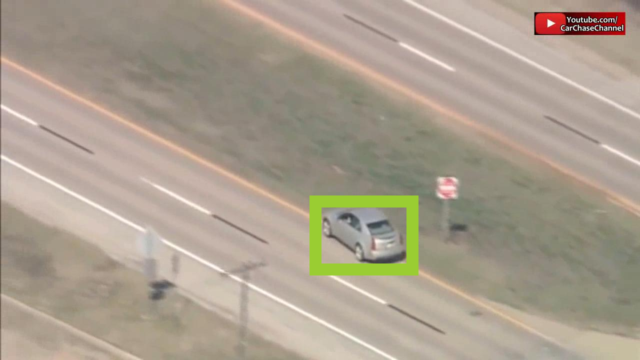} &
\includegraphics[width=\fw]{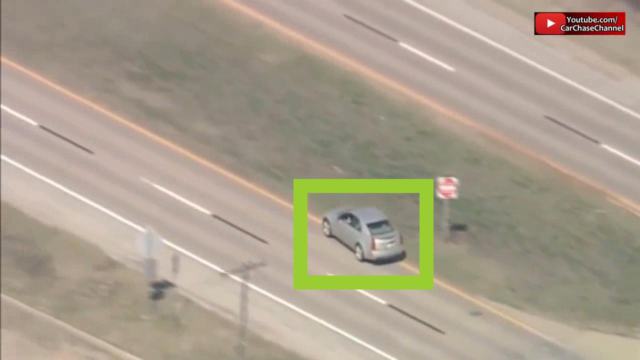} &
\includegraphics[width=\fw]{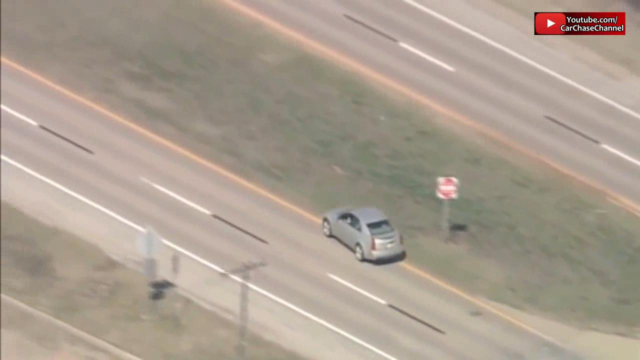} &
\includegraphics[width=\fw]{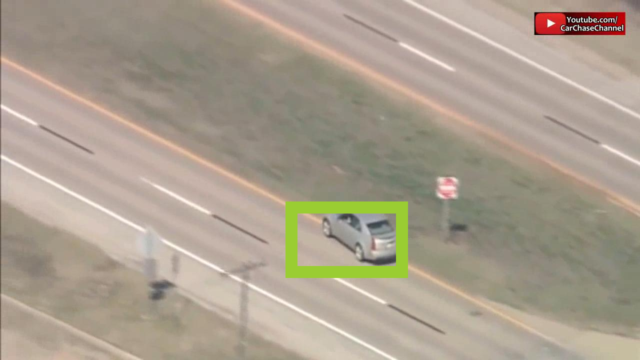} &
\includegraphics[width=\fw]{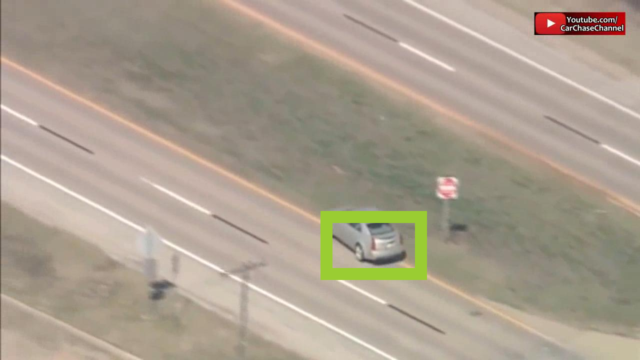} \\
&
\rotatebox{90}{\parbox{10mm}{\centering  \vspace{3pt} \quad \tiny Negative}} &
\includegraphics[width=\fw]{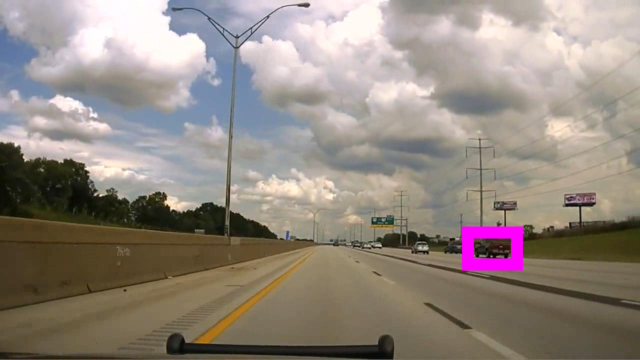} &
\includegraphics[width=\fw]{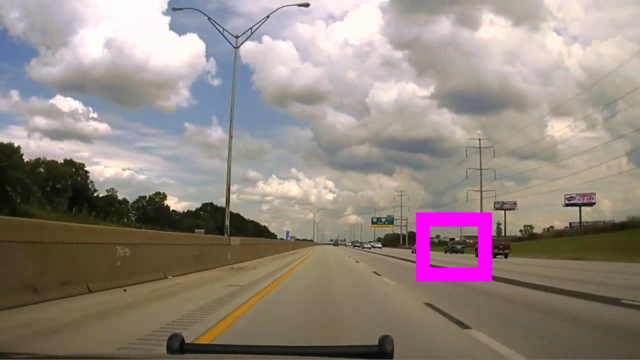} &
\includegraphics[width=\fw]{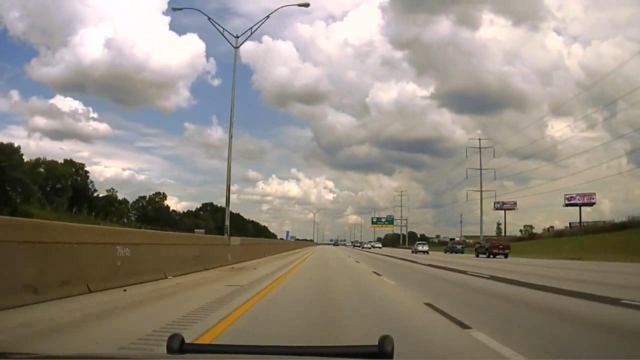} &
\includegraphics[width=\fw]{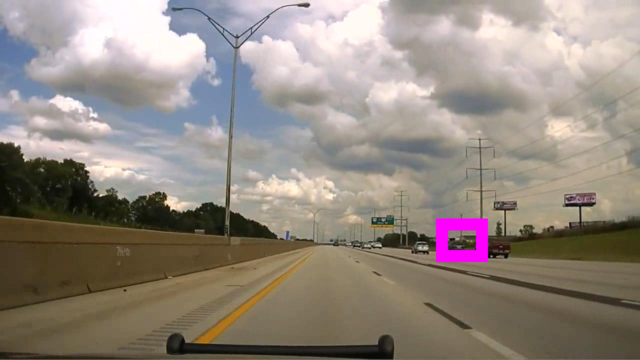} &
\includegraphics[width=\fw]{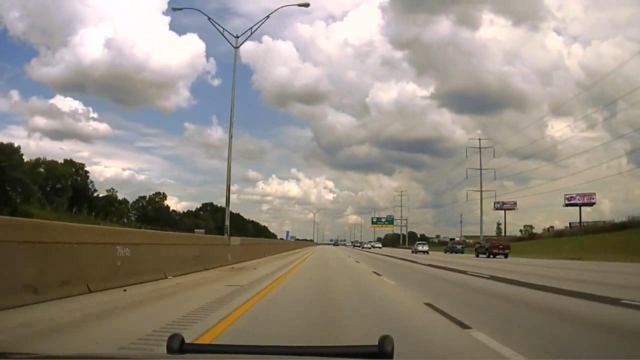} \\
\end{tabular}
\caption{\textbf{[In-context inference for personalized object identification and localization task]}
(Left) Examples of reference data, (Right) positive and negative query images, and inference results using Florence-2, No-Time-To-Train (NT3), Qwen2-VL with prompting, IPLoc, and the proposed IPLoc-ID, respectively: Red boxes indicate reference annotations, green boxes indicate correct detections, and magenta boxes indicate false-positive detections. More detail is shown in Section~\ref{sec:results}.}
\label{fig:incontext-inference-for-POIL}
\end{figure}
%
%
%
\par
To further unlock the object detection capability of VLMs for practical applications, 
we revisit the localization setting of IPLoc and introduce a more general task, termed personalized object identification and localization (POIL).
In this task, the model is required to output a bounding box when the same object instance specified by the reference data exists in the query image, and to reject the query image otherwise. 
Figure~\ref{fig:incontext-inference-for-POIL} illustrates the characteristic behavior of existing methods under the POIL setting.
Conventional OD and FSOD methods operate mainly at the category level and may therefore produce false-positive detections on negative examples containing object instances different from the reference object.
Similarly, IPLoc has no explicit mechanism for rejecting negative query images and exhibits the same failure mode.
We also construct customized datasets based on public video object tracking datasets.
The constructed datasets consist of instance-level positive and negative examples and are suitable for fine-tuning and evaluating models under the POIL task.
\par
We further propose an in-context algorithm, IPLoc-ID, for solving the POIL task with VLMs. 
IPLoc-ID reinterprets the BBOX, which is treated as the final output in IPLoc, as a reference-conditioned candidate, and then determines whether this candidate actually corresponds to the reference object instance through a subsequent identification component. 
Specifically, based on the autoregressive generation process of VLMs, IPLoc-ID generates the BBOX, a self-posed query, and an identification answer as a single continuous text sequence. 
The self-posed query connects the input reference data, the query image, and the previously generated BBOX into a single token sequence, naturally eliciting the final Yes/No identification response. 
By fine-tuning with both positive and negative examples, IPLoc-ID jointly learns reference-conditioned candidate localization and instance-level identification. 
Through ablation studies and comprehensive experiments on four datasets, we show that IPLoc-ID substantially suppresses false-positive detections on negative query images while maintaining the localization performance of IPLoc. 
As a result, the proposed method shows a clear advantage in instance-level identification and localization, which cannot be sufficiently addressed by conventional OD, FSOD methods, or the previous localization-only IPLoc framework.
\par
The remainder of this paper is organized as follows. 
Section~\ref{sec:related-works} summarizes related works. 
Section~\ref{sec:method} describes the proposed framework. 
Section~\ref{sec:results} reports empirical results. 
Section~\ref{sec:conclusion} concludes this paper.
%
%
%
%
%
%

%
%
%
%
%
%
%
%
%
%
%
\section{Related Works} \label{sec:related-works}
\textbf{Few-shot object detection (FSOD)} aims to detect novel object categories from a few annotated examples~\cite{kohler2023few,xin2024few}. 
Representative approaches include fine-tuning-based methods such as TFA~\cite{wang2020frustratingly}, FSCE~\cite{sun2021fsce}, and DeFRCN~\cite{qiao2021defrcn}, meta-learning or support-query interaction methods such as Meta R-CNN~\cite{yan2019meta} and fully cross-transformer-based methods~\cite{han2022few}, and recent transformer- or foundation-model-based methods such as DE-ViT~\cite{zhang2024devit}.
More recent FSOD methods exploit vision foundation models: FT-FSOD~\cite{yu2026acloser} fine-tunes Grounding-DINO~\cite{liu2024groundingdino} with support data, FSOD-VFM (VFM)~\cite{feng2026fsodvfm} constructs class-wise prototypes using UPN, SAM2, and DINOv2, and No-Time-To-Train~\cite{espinosa2025notime} performs training-free matching between support features and SAM-generated masks.
The key difference from our POIL task is that FSOD is category-level, whereas POIL is instance-level.
In FSOD, detecting any object of the target category is sufficient.
In contrast, POIL requires localizing the specific reference-conditioned instance and rejecting other instances, including in-class distractors.
Moreover, while recent FSOD pipelines often rely on task-specific modules such as SAM, IPLoc variants aim to exploit the general-purpose visual reasoning ability of VLMs.
In our experiments, we compare with VFM and No-Time-To-Train to highlight this difference between category-level FSOD and instance-level POIL.
\par
\textbf{Instance-level retrieval and localization.}
Recent studies have extended retrieval and localization toward instance-level matching.
i-CIR~\cite{psomas2025instance} retrieves a specific instance using a visual query and textual modification, but does not perform spatial localization.
REIR~\cite{hao2025referring} retrieves and localizes instances using fine-grained natural language expressions, whereas POIL specifies the target through reference images, labels, and bounding boxes.
Few-Shot Object Localization (FSOL)~\cite{ren2024fsol} localizes objects from limited support examples, but does not explicitly address negative query rejection.
These studies show the importance of instance-level retrieval and localization.
In contrast to them, POIL focuses on VLM-based in-context inference, where the reference-conditioned instance must be localized when present and rejected when absent.
\par
\textbf{Vision-language models.}
Vision foundation models have become central to computer vision.
Contrastive vision-language models such as CLIP~\cite{radford2021clip} and OpenCLIP~\cite{cherti2023openclip} learn joint image-text representations, while BLIP variants~\cite{li2022blip,li2023blip2} combine visual understanding with language generation.
Task-specific foundation models such as SAM~\cite{ravi2025sam} and DINOv2~\cite{oquab2023dinov2} provide visual modules for downstream pipelines.
VLMs, including LLaVA~\cite{liu2023llava}, Gemma~\cite{gemmateam2025gemma3}, and Qwen-VL~\cite{Qwen2-VL,Qwen3-VL}, integrate visual perception and language generation, enabling instruction-based visual reasoning, detection, and grounding~\cite{liu2023llava,Qwen2-VL,Qwen3-VL,zhang2024llava-grounding,yao2026qwen3seg}.
For localization tasks, an important challenge is to design prompts and input contexts that make VLMs produce reliable structured outputs such as bounding boxes.
IPLoc~\cite{doveh2025iploc} addresses this direction by formulating personalized object localization as sequence generation.
Our work extends IPLoc by adding an identification component based on a self-posed query, enabling the model to decide whether the localized candidate matches the reference instance.
\par
\vspace{8pt} 
\textbf{Self-posed query.}
Question generation and self-questioning have been studied as mechanisms for improving model reasoning in language tasks~\cite{press2023measuring,qi2023socratic} and vision-language tasks~\cite{sun2024sqllava,prasad2024rephrase}. 
Our self-posed query is inspired by this idea but differs in purpose and formulation.
Rather than generating diverse or recursive questions for general reasoning, IPLoc-ID uses a fixed intermediate query to connect the generated BBOX candidate with the final identification answer.
This design induces a simple sequence from context, to localization, to identification, and enables instance-level rejection within VLM-based inference.
\par
\textbf{In-context learning and personalized object localization.}
In-context learning allows a model to solve a task using examples and context provided in the input, without updating model parameters at test time~\cite{min2022metaicl}.
Although originally studied in language models, it has also been extended to VLM settings~\cite{monajatipoor2023metavl,yu2024eliciting,sheng2024towards}.
Here, the model may be trained beforehand to acquire the task format, but the reference data is used only as input context in inference.
IPLoc~\cite{doveh2025iploc} is a representative in-context approach for personalized object localization.
Our work follows this paradigm and extends it to a more practical setting where query images may or may not contain the intended object instance.
%
%
%

%
%
%
%
%
%
%
\newcommand{\refimagek}{I^{\text{r}}_k}
\newcommand{\reflabel}{\ell}
\newcommand{\refbboxk}{B^{\text{r}}_k}
\newcommand{\queimage}{I^{\text{t}}}
\newcommand{\quelabel}{\ell}
\newcommand{\tarbbox}{B^{\text{t}}}
\newcommand{\ans}{A}
\def\iplocbbox{B}
\def\iplocoutput{y}
\section{Method} \label{sec:method}
This section defines POIL and formulates IPLoc and its limitation under this setting.
We then present IPLoc-ID as a VLM-based in-context solution and describe the customized POIL datasets.
\subsection{Preliminary} \label{sec:method:preliminary}
\subsubsection{Personalized object identification and localization}
We first introduce personalized object identification and localization (POIL).
Here, a ``personalized object''~\cite{doveh2025iploc} denotes a specific object instance specified by reference data, such as reference images, labels, or annotations.
Given such reference data and a target image, POIL aims to identify and localize the same object instance in the target image.
\par
Conceptually, POIL extends POL by introducing negative-query rejection and can also be viewed as an instance-level counterpart of conventional FSOD.
While FSOD detects objects at the category level from support data, POIL requires detecting a specific object instance.
Although the original IPLoc also localizes a specific object from the same input data, its formulation can produce false positives for objects other than the intended instance.
In POIL, the model must localize the target object only when the same instance as the reference object appears in the target image; otherwise, it must reject the image.
This property is essential for instance-level applications such as image retrieval, video grounding, and object identification.
\par
Formally, we define the POIL task as follows.
Let the input be defined as
%
\begin{equation} \label{eq:incontext-input}
x = \{(\refimagek, \reflabel, \refbboxk)\}_{k=1}^{N}, \; \queimage, \; \quelabel,
\end{equation}
%
where $\refimagek$ denotes the $k$-th reference image, $\refbboxk$ denotes its annotated bounding box, and $\reflabel$ denotes the corresponding class label. 
$\queimage$ and $\quelabel$ denote the target query image and its query label, respectively. 
This input format follows the original IPLoc formulation. 
The input $x$ is converted into a sequence of tokens and fed into a transformer-based VLM.
\par
Let $\mathcal{X}$ be the input space, where each input 
$x \in \mathcal{X}$ consists of reference data and a query image. 
Importantly, different from the original IPLoc, we do not restrict $\mathcal{X}$ to inputs in which the query image necessarily contains the object specified by the reference data. 
This allows practical scenarios where the query image $\queimage$ does not contain the same object instance as the reference data. 
We refer to such inputs as ``negative examples'', in contrast to ``positive examples'' whose query image contains the object of interest.
To handle both positive and negative data, we define the identification condition
%
\begin{equation}
\delta(x) \in \{0,1\},
\end{equation}
%
where $\delta(x)=1$ if the query image in $x$ contains the same object instance as specified by the reference data, and $\delta(x)=0$ otherwise.
\par
Let $\mathcal{B}$ denote the bounding-box space and $\varnothing$ denote rejection.
We then define the ideal task of POIL as a mapping
%
\begin{equation} \label{eq:oil-task-mapping}
f^\ast : \mathcal{X} \rightarrow \mathcal{B} \cup \{\varnothing\},
\end{equation}
%
such that
%
\begin{equation} \label{eq:oil-task-fx}
f^\ast(x) =
\begin{cases}
\tarbbox, & \text{if } \delta(x)=1,\\
\varnothing, & \text{if } \delta(x)=0.
\end{cases}
\end{equation}
%
where $\tarbbox \in \mathcal{B}$ denotes the ground-truth bounding box of the reference object instance in the query image when it exists.
Our main objective is to develop an algorithm that accurately approximates the ideal mapping in Eq.~(\ref{eq:oil-task-fx}).
\subsubsection{Evaluation metric for POIL}
Following common OD and FSOD evaluation metrics, we use mIoU~\cite{everingham2010pascal,lin2014coco} to measure BBOX localization accuracy and F1-score~\cite{powers2011evaluation} to evaluate instance-level identification, including false-positive suppression on negative query images.
\subsection{The baseline IPLoc} \label{sec:method:iploc}
\subsubsection{Formulation of IPLoc}
IPLoc~\cite{doveh2025iploc} is an in-context algorithm for personalized object localization (POL).
Following the input format $x$ defined in Eq.~(\ref{eq:incontext-input}), POL aims to generate the bounding-box coordinates of the object category or instance specified by the reference data.
Since IPLoc is trained with the standard next-token-prediction objective to generate localization coordinates, its VLM-based output can be formulated as conditional generation of the target bounding box.
Specifically, given an input $x$, the output sequence of IPLoc can be written as
%
\begin{equation} \label{eq:iploc:generated-y}
\iplocoutput
=
\langle \iplocbbox \rangle,
\end{equation}
%
where $\iplocbbox\in \mathcal{B}$ denotes the estimated bounding-box coordinates in the query image, and $\langle \cdot \rangle$ denotes a generated text component in the output sequence.
Equivalently, IPLoc parameterized by $\theta$ models the conditional probability 
%
\begin{equation}
p_{\theta}(\tarbbox \mid x),
\end{equation}
%
and generates the BBOX component as 
%
\begin{equation}
\iplocbbox = \arg\max_{b \in \mathcal{B}} p_{\theta}(b \mid x).
\end{equation}
%
%
%
%
\par 
During fine-tuning, IPLoc constructs multi-modal conversations from image sequences. 
Each conversation consists of the input $x$ and the ground-truth BBOX for the query image as the target output. 
Accordingly, IPLoc is trained by minimizing the negative log-likelihood of the target bounding box:
%
\begin{equation} \label{eq:iploc:objective}
\min_{\theta}
\mathbb{E}_{(x,\tarbbox)\sim \tilde{P}}
\left[
-\log p_{\theta}(\tarbbox \mid x)
\right],
\end{equation}
%
where $\tilde{P}$ denotes the POL training distribution, in which the query image contains the object specified by the reference data.
The original IPLoc also introduces pseudo-label-based label noise. 
Specifically, the class label $\reflabel$ in the input sequence is randomly replaced with a pseudo label during training. 
This reduces overfitting to specific class names and encourages localization based on visual examples and bounding-box annotations.
\par
A key insight of IPLoc is to exploit the contextual understanding ability of transformer-based VLMs by formulating localization as sequence generation. 
In this formulation, the personalized examples establish a repeated order of image, label, and bounding-box coordinates. 
Given a query with only an image and label, this format induces the model to complete the sequence by predicting the missing bounding box.
\subsubsection{Limitation of IPLoc}
However, IPLoc assumes that the object of interest is always present in the query image.
This limitation follows from its output-space constraint: IPLoc maps any input $x \in \mathcal{X}$ to a bounding box in $\mathcal{B}$.
In contrast, for a negative query image satisfying $\delta(x)=0$, the ideal POIL mapping in Eq.~(\ref{eq:oil-task-fx}) requires the rejection output $f^\ast(x)=\varnothing$.
Since $\varnothing \notin \mathcal{B}$, IPLoc cannot represent the ideal output for negative query images.
Equivalently, even when the query image does not contain the reference object instance, IPLoc still returns a bounding box:
%
\begin{equation} \label{eq:iploc:f-for-negative}
f_{\mathrm{IPLoc}}(x) \in \mathcal{B}
\quad \text{even when} \quad
f^\ast(x)=\varnothing.
\end{equation}
%
This mismatch between the IPLoc formulation in Eq.~(\ref{eq:iploc:f-for-negative}) and the ideal POIL mapping in Eq.~(\ref{eq:oil-task-fx}) leads to false-positive detections on negative query images.
\par 
This false-positive behavior, as discussed in Section~\ref{sec:introduction}, limits the applicability of IPLoc under the POIL setting, where target images may or may not contain the reference object. Thus, the model must localize true-positive cases while also identifying true-negative cases and rejecting query images without the reference object. To address this requirement, we extend IPLoc to IPLoc-ID, which incorporates identification into personalized object localization. %
%
%
%
%
%
%
%
%
%
%
\def\iplocidbbox{B}
\def\iplocidquery{Q}
\def\iplocidanswer{A}
\def\iplocidtargetanswer{A^\ast}
\def\iplocidoutput{y}
\def\iplocidmodel{f}
%
\subsection{The Proposed IPLoc-ID}  \label{sec:method:iplocid}
\begin{figure}[htb]
\centering
\includegraphics[width=\textwidth]{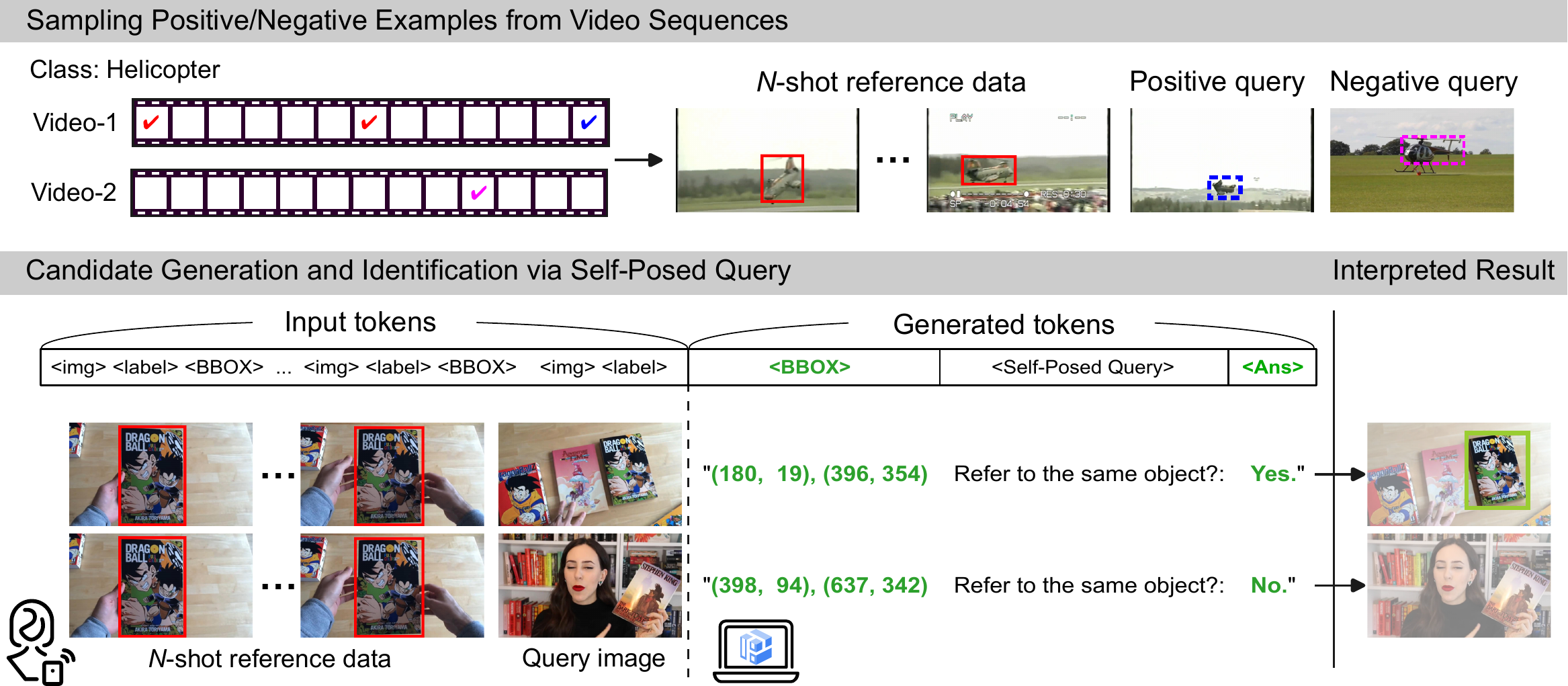}
\caption{\textbf{[The proposed IPLoc-ID framework]}
(Top) We introduce personalized object identification and localization (POIL) and construct datasets by augmenting video object tracking data with negative query images.
(Bottom) IPLoc-ID extends the sequence-generation formulation of IPLoc by generating a BBOX candidate, a self-posed query, and an identification answer in an autoregressive process, enabling the model to reject negative query images.}
\label{fig:overview}
\end{figure}

We now propose IPLoc-ID as an in-context algorithm for solving POIL.
IPLoc-ID leverages the strong generalization ability of VLMs to localize the object of interest in positive query images while rejecting negative query images that do not contain the reference object instance. 
To this end, we decompose Eq.~(\ref{eq:oil-task-fx}) into BBOX localization and identification, and generate them sequentially as text using a VLM.
The BBOX component first produces a candidate bounding box in the query image, following the sequence-generation principle of IPLoc. 
The identification component then verifies whether the generated BBOX corresponds to the object instance specified by the reference data. 
To connect these components, we introduce a self-posed query, which preserves the natural sequence from input data to generated text.
Finally, an interpreter function converts the generated text into the structured output required by POIL.
Figure~\ref{fig:overview} illustrates the overall framework of IPLoc-ID.
%
%
%
%
%
%
%
%
%
%
%
%
%

%
%
%
\subsubsection{Sequential generation of localization and identification}
IPLoc-ID generates the following text sequence:
%
\begin{equation} \label{eq:iplocid:generated-y}
\iplocidoutput
=
\langle \iplocidbbox \rangle
\;
\langle \iplocidquery \rangle
\;
\langle \iplocidanswer \rangle,
\end{equation}
%
where $\iplocidbbox \in \mathcal{B}$ denotes the generated bounding box, 
$\iplocidquery$ denotes the fixed self-posed query, and 
$\iplocidanswer \in \mathcal{A}$ denotes the identification answer, with 
$\mathcal{A}=\{\mathrm{Yes}, \mathrm{No}\}$.
The answer $\iplocidanswer$ is designed to approximate the identification condition $\delta(x)$:
%
\begin{equation}
\iplocidtargetanswer(x)
=
\begin{cases}
\mathrm{Yes}, & \text{if } \delta(x)=1,\\
\mathrm{No}, & \text{if } \delta(x)=0.
\end{cases}
\end{equation}
%
Thus, the BBOX component provides a candidate localization, while the answer component determines whether the candidate should be accepted or rejected.
\par
The autoregressive generation process of IPLoc-ID using a VLM parameterized by $\theta$ can be factorized as
%
\begin{equation}
p_{\theta}(\iplocidoutput \mid x)
=
p_{\theta}(\iplocidbbox \mid x) \cdot
\;
p_{\theta}(\iplocidquery \mid x, \iplocidbbox) \cdot
\;
p_{\theta}(\iplocidanswer \mid x, \iplocidbbox, \iplocidquery).
\end{equation}
%
Since $\iplocidquery$ is fixed by the output format, 
$p_{\theta}(\iplocidquery \mid x, \iplocidbbox)=1$.
Thus, IPLoc-ID effectively decomposes output generation into BBOX generation and identification:
%
\begin{equation}
p_{\theta}(\iplocidoutput \mid x)
=
p_{\theta}(\iplocidbbox \mid x)
\; \cdot
p_{\theta}(\iplocidanswer \mid x, \iplocidbbox, \iplocidquery).
\end{equation}
%
\par
During inference, IPLoc-ID first generates the BBOX component:
%
\begin{equation} \label{eq:iplocid:y1}
y_1
=
\langle \iplocidbbox \rangle,
\qquad
\iplocidbbox
=
\arg\max_{b \in \mathcal{B}}
p_{\theta}(b \mid x).
\end{equation}
%
This follows the same BBOX-generation process as IPLoc, but produces a reference-conditioned candidate bounding box in the query image.
Second, IPLoc-ID appends the fixed self-posed query:
%
\begin{equation} \label{eq:iplocid:y2}
y_2
=
\langle \iplocidquery \rangle.
\end{equation}
%
Unlike prior self-questioning methods~\cite{press2023measuring,qi2023socratic,sun2024sqllava}, our fixed self-posed query does not introduce an additional stochastic decision, but bridges the generated BBOX candidate and the final identification answer.
The effect of this design is analyzed in Section~\ref{sec:results:ablation}.
\par
Third, IPLoc-ID generates the identification answer:
%
\begin{equation} \label{eq:iplocid:y3}
y_3
=
\langle \iplocidanswer \rangle,
\qquad
\iplocidanswer
=
\arg\max_{a \in \mathcal{A}}
p_{\theta}(a \mid x,\iplocidbbox,\iplocidquery),
\end{equation}
%
where $\mathcal{A}=\{\mathrm{Yes},\mathrm{No}\}$ denotes the answer space.
The final output in Eq.~(\ref{eq:iplocid:generated-y}) is then formed by the sequential generation of $y_1$, $y_2$, and $y_3$.
\subsubsection{Fine-tuning with positive and negative examples}
In-context learning allows a model to solve a task using reference examples and contextual information in the input, without updating model parameters at test time~\cite{min2022metaicl}. 
Following this principle, IPLoc-ID performs inference without updating the model on test data. 
To acquire the required task format and contextual reasoning ability, we fine-tune the VLM for POIL using both positive and negative examples.
\par
The fine-tuning objective of IPLoc-ID is formulated as
$\min_{\theta} F(\theta)$, where $\theta$ denotes the model parameters and $F(\theta)$ maximizes the likelihood of the entire target output sequence. 
Specifically, $F(\theta)$ is defined as
%
\begin{equation} \label{eq:iplocid:objective-f}
\begin{aligned}
F(\theta) :=
\mathbb{E}_{(x,\iplocidoutput^\ast)\sim P}
\left[
-\log p_{\theta}(\tarbbox \mid x)
-\log p_{\theta}(\iplocidquery \mid x, \tarbbox)
-\log p_{\theta}(\iplocidtargetanswer(x) \mid x, \tarbbox, \iplocidquery)
\right],
\end{aligned}
\end{equation}
%
where $P$ denotes the POIL training distribution, including both positive and negative examples. 
The target output sequence $\iplocidoutput^\ast$ is defined as
%
\begin{equation} \label{eq:iplocid:target-sequence}
\iplocidoutput^\ast
=
\langle \tarbbox \rangle
\;
\langle \iplocidquery \rangle
\;
\langle \iplocidtargetanswer(x) \rangle,
\end{equation}
%
where $\tarbbox$ denotes the target BBOX, $\iplocidquery$ denotes the fixed self-posed query, 
and $\iplocidtargetanswer(x)$ denotes the ground-truth identification answer determined by whether the query image contains the reference object instance.
\par
Positive examples contain the same object instance as specified by the reference data, and their target answer is $\mathrm{Yes}$.
Negative examples use query images that do not contain the reference object instance, and their target answer is $\mathrm{No}$.
To learn reference-conditioned instance-level discrimination, we construct negative examples from different object instances, including instances from the same category as the positive examples.
More details of dataset construction are provided in Section~\ref{sec:method:dataset}.
\par
For positive examples, $\tarbbox$ is the ground-truth bounding box of the reference object instance in the query image.
For negative examples, $\tarbbox$ is the BBOX of the most plausible candidate object in the query image.
Thus, even for negative examples, the model first generates a candidate BBOX following the same sequence-generation process.
However, this BBOX is treated only as a reference-conditioned candidate region, and the subsequent identification answer determines whether it corresponds to the reference object instance.
This enables the model to learn both candidate localization and instance-level rejection.
\subsubsection{Interpretation of generated results}
We employ an interpreter $\gamma(y)$ to extract the generated BBOX text $\langle \iplocidbbox \rangle$ and answer text $\langle \iplocidanswer \rangle$ from the generated text.
The implementation of the interpreter $\gamma$ is described in Section~\ref{sec:method:implementation}.
We define the interpreter output as
%
\begin{equation} 
\gamma(\iplocidoutput) =
\begin{cases}
1, & \text{if } \langle \iplocidanswer \rangle \text{ is interpreted as positive},\\
0, & \text{if } \langle \iplocidanswer \rangle \text{ is interpreted as negative}.
\end{cases}
\end{equation}
%
Based on this interpreter output, we define the final prediction as
%
\begin{equation} \label{eq:iplocid:interpreted}
\iplocidmodel(x)
=
\begin{cases}
\iplocidbbox, & \text{if } \gamma(\iplocidoutput)=1,\\
\varnothing, & \text{if } \gamma(\iplocidoutput)=0,
\end{cases}
\end{equation}
%
where $\iplocidbbox$ denotes the generated BBOX coordinates. 
This IPLoc-ID formulation approximates the ideal mapping in Eq.~(\ref{eq:oil-task-fx}) by returning a BBOX for positive query images and rejecting negative query images.
%
%
%

%
%
%
%
%
\begin{table}[htb]
\centering
\caption{Examples of generated texts and their interpreted model response.}
\label{tab:interpreter}
\renewcommand{\arraystretch}{1.4}
\scriptsize
\begin{tabular}{m{0.80\linewidth} >{\centering\arraybackslash}m{0.125\linewidth}}
\Xhline{1.2pt}
Generated text &  Response \\
\hline
\textit{``[175.9, 411.1, 656.3, 866.7]''}
& positive \\
\textit{``Not found.''}
& negative \\
\textit{``bbox=[197.0, 388.0, 640.0, 843.0]\textbackslash n same\_object=NO''}
& negative \\
\textit{``[382.1, 233.3, 595.8, 605.6], Do all these boxes have the same object? Yes.''} 
& positive\\
\textit{``[187.5, 405.6, 656.3, 855.6], Do all these boxes have the same object? No.''} 
& negative \\
\Xhline{1.2pt}
\end{tabular}
\end{table}
%
%
%
%
\subsection{Implementation of IPLoc-ID} \label{sec:method:implementation}
\par
\textbf{Interpreter function.}
Since VLMs generate free-form text, an interpreter is required to convert generated text into structured outputs.
In this study, the interpreter extracts bounding-box coordinates ($\langle$BBOX$\rangle$) and the identification response ($\langle$Ans$\rangle$).
If the generated text contains an explicit negative expression such as \texttt{No}, \texttt{Not found}, \texttt{different}, or \texttt{not the same}, or contains no valid bounding box, it is classified as a \textit{negative response}.
Otherwise, it is treated as a \textit{positive response}, allowing methods without an identification component, such as the original IPLoc, to be consistently interpreted as positive.
Table~\ref{tab:interpreter} shows examples of model outputs and their interpreted responses.
\par
\textbf{LoRA fine-tuning.}
For fine-tuning, we employ LoRA~\cite{hu2022lora}, a standard parameter-efficient fine-tuning method.
Let $\theta_0$ denote the frozen parameters, and let $\phi$ denote the trainable LoRA parameters.
The effective model parameters are written as
$\theta(\phi)=\theta_0+\Delta_{\mathrm{LoRA}}(\phi)$, 
where $\Delta_{\mathrm{LoRA}}(\phi)$ denotes the low-rank update.
Accordingly, we optimize Eq.~(\ref{eq:iplocid:objective-f}) with respect to $\phi$ as
$\min_{\phi} F\left(\theta(\phi)\right)$.
\subsection{Datasets for the POIL task} \label{sec:method:dataset}
We construct datasets for fine-tuning and evaluating the POIL task.
We use four public sources: LaSOT~\cite{fan2019lasot}, PDM (Burst)~\cite{samuel2024pdm}, GOT-10K~\cite{huang2019got}, and VastTrack~\cite{peng2024vasttrack}.
These datasets are selected because they \textup{(i)} consist of image sequences, 
\textup{(ii)} provide annotations suitable for BBOX localization, and 
\textup{(iii)} include class labels.
Among them, LaSOT provides high data volume and class diversity, with multiple sub-classes per class corresponding to different object instances or video sequences.
We therefore use a subset of LaSOT for fine-tuning, and use the LaSOT test split and the other datasets for evaluation.
\par
Table~\ref{tab:datasets} summarizes the customized datasets.
Each data sample contains one positive and one negative query image that share the same reference data.
Thus, the actual number of examples is twice the number of data samples.
For example, the LaSOT training set contains 700 data samples, corresponding to 700 positive and 700 negative examples.
For simplicity, we refer to each customized dataset by its source name.
\subsubsection{Sampling procedure}
The sampling procedure is illustrated in the top part of Figure~\ref{fig:overview}.
Reference data and positive query images are sampled following the original IPLoc, while negative query images are newly introduced for POIL.
From each video sequence, we uniformly sample $N+1$ frames: the first $N$ frames are used as reference data, and the last frame is used as the \textit{positive query image} with its ground-truth BBOX.
We then sample one \textit{negative query image} from a different instance, either from a different class or from a different sub-class within the same class.
The positive and negative query images share the same reference data but are treated as independent query cases during training and evaluation.

\subsubsection{Training set}
We construct the training set from LaSOT.
The public LaSOT dataset contains approximately 70 classes, and each class contains multiple sub-classes corresponding to different object instances or video sequences.
\par
We split the classes into training and test splits and apply the above sampling procedure to the training split.
Negative query images are sampled from different sub-classes within the same class, providing in-class adversarial examples.
As a result, we obtain 700 training samples, each consisting of $N$ reference images, one positive query image, and one negative query image.

\subsubsection{Test set}
We construct 140 LaSOT test samples from the held-out class split, using in-class negative query images.
This test set evaluates generalization to unseen classes within the same domain.
\par
For unseen-domain evaluation, we use PDM, GOT-10K, and VastTrack without fine-tuning on these datasets.
For PDM and GOT-10K, which do not provide explicit sub-class structures, negative query images are sampled from different classes.
For VastTrack, we select approximately 400 classes with at least two sub-classes and sample in-class negative query images from different sub-classes within the same class.
Thus, VastTrack provides a larger unseen-domain test set with more challenging in-class negative examples.

\subsubsection{$N$-shot settings}
We evaluate four $N$-shot settings: $N = 1, 2, 4, 8$.
To isolate the effect of $N$, datasets with smaller $N$ are constructed as subsets of the $N=8$ set, sharing the same positive and negative query images.
Because IPLoc-ID is sensitive to the number of reference images used during fine-tuning, we fine-tune separate models for each $N$ and evaluate them on the corresponding $N$-shot test sets.
%
%
%

%
%
%
\def \pw {66pt}
\def \qw {50pt}
\begin{table}[htb] 
\centering
\caption{Summary of customized datasets.}
\label{tab:datasets}
%
%
\small
%
%
\begin{tabular}{p{\pw} | P{\qw}P{\qw}p{\qw}p{\pw}}
\Xhline{1.2pt}
Dataset & \#Training & \#Test & $N$-shot & Negative data \\
\hline
LaSOT~\cite{fan2019lasot}     & 700 & 140 & 1, 2, 4, 8 & in-class  \\
PDM~\cite{samuel2024pdm}       & --  & 745 & 1, 2 & out-of-class \\
GOT-10K~\cite{huang2019got}   & --  & 180 & 1, 2, 4, 8 & out-of-class \\
VastTrack~\cite{peng2024vasttrack} & --  & 400 & 1, 2, 4, 8 & in-class  \\
\Xhline{1.2pt}
\end{tabular}
%
%
\end{table}
%
%

%
%
%
%
%
%
%
%
\section{Experimental Results} \label{sec:results}
In this section, we first present the experimental setup and discuss the selection of backbone models.
Then, we conduct ablation studies. Finally, we report the main results.
\subsection{Experimental Setup} \label{sec:results:setup}
\subsubsection{Training procedure}
Using LoRA, we fine-tune each backbone model on the customized training set.
For each backbone, we train two variants: IPLoc using Eq.~(\ref{eq:iploc:objective}) and IPLoc-ID using Eq.~(\ref{eq:iplocid:objective-f}).
During training, positive and in-class negative pairs are sequentially fed in a randomized order, enabling IPLoc-ID to learn instance-level decision boundaries.
The same pairs are also used to train IPLoc, ensuring that IPLoc and IPLoc-ID are trained on identical data.
This does not change the formulation of IPLoc, which performs only BBOX localization for each target image.
We also observe that using negative examples improves IPLoc mIoU, suggesting their role as data augmentation even for localization-only training.
\par
Our reproduced IPLoc may not be strictly identical to the original IPLoc, as its complete training configuration and scripts are not publicly available.
Moreover, the original IPLoc does not consider negative examples and was trained using three datasets~\cite{doveh2025iploc}.
Nevertheless, our reproduced IPLoc shows the expected localization-only behavior under standard training settings.
We also include the partially released official Qwen2-VL-7B IPLoc model, denoted by ``IPLoc 7B (official)'', as a reference in the final comparison.
%
%
\subsubsection{Evaluation procedure}
The final model is further evaluated on the test datasets using mIoU and F1-score.
mIoU measures BBOX localization accuracy, while F1-score assesses identification performance.
Since positive and negative query images are balanced in our setting, methods that always return positive responses, such as localization-only detectors and IPLoc, yield an F1-score of $2/3$ ($\simeq 0.667$).
Thus, this value serves as the theoretical baseline for methods without an explicit negative-query rejection mechanism.
Unless otherwise specified, we report the average metrics over three independent training/evaluation runs.

\subsubsection{Other experimental details}
Our implementation is based on Hugging Face, and all backbone models are publicly available on the same platform.
We use LoRA with rank $r=8$ and scaling factor $\alpha=16$, following common LoRA fine-tuning settings and the publicly disclosed details of the original IPLoc configuration.
The remaining hyperparameters follow standard Hugging Face LoRA fine-tuning examples.
Model training and inference were primarily conducted on four NVIDIA A100 GPUs.
For Qwen3-VL-235B, we used eight NVIDIA B200 GPUs.
\subsection{Backbone Model Selection} \label{sec:results:backbone} 
The proposed method assumes a transformer-based VLM with autoregressive text generation. We empirically select the backbone architecture and model size based on performance as follows.
\subsubsection{Model architecture}
We first compare the following VLMs: LLaVA1.5-7B, Gemma3-12B, Qwen2-VL-7B, and Qwen3-VL-8B, as shown in Table~\ref{tab:ablation:backbone}. 
These models are representative open-source pretrained VLMs. 
LLaVA1.5-7B is one of the early instruction-tuned models.
Gemma3-12B is a recent VLM with strong conversational ability.
Qwen2-VL-7B is one of the backbones used in the previous IPLoc study. 
Qwen3-VL-8B represents the next generation of the Qwen series, and we further examine its larger variants.
\par
Table~\ref{tab:ablation:backbone} shows that the Qwen models outperform Gemma3 and LLaVA1.5 in mIoU. 
One possible explanation is that the Qwen series is more effective for visual localization and structured coordinate generation in our setting. 
In addition, Qwen3-VL-8B outperforms Qwen2-VL-7B in both mIoU and F1-score under similar model sizes.
Based on these results, we adopt Qwen3-VL-8B as the main backbone of our method, while Qwen2-VL-7B is also included in the basic analyses for consistency with the previous IPLoc study.
\subsubsection{Model-size scalability}
We compare three Qwen3-VL variants: 8B, 32B, and 235B, as shown in Table~\ref{tab:ablation:backbone}. 
Qwen3-VL-8B and Qwen3-VL-32B are dense models, whereas Qwen3-VL-235B adopts a Mixture-of-Experts (MoE) architecture, where tokens are dynamically routed to a subset of experts.
As shown in Table~\ref{tab:ablation:backbone}, performance improves with model scale, and the result with Qwen3-VL-235B suggests that IPLoc-ID can also benefit from the known scalability of sparse MoE models~\cite{riquelme2021scaling}.
\par
In the following ablation studies, we mainly use Qwen3-VL-8B and Qwen3-VL-32B as representative dense backbones for controlled experiments, and include Qwen3-VL-235B in the final comprehensive evaluation.
\def \pw{23pt}
\begin{table}[t]
\centering
\caption{\textbf{[Backbone model selection]} mIoU and F1-score on the LaSOT test set for backbones under different $N$-shot settings.}
\label{tab:ablation:backbone}
\scriptsize
\vspace{6pt}
%
%
\begin{tabular}{l|P{\pw}P{\pw}P{\pw}P{\pw}|P{\pw}P{\pw}P{\pw}P{\pw}}
\Xhline{1.2pt}
\multicolumn{1}{c|}{}
& \multicolumn{8}{c}{mIoU ($\uparrow$)} \\
\cline{2-9}
& \multicolumn{4}{c|}{IPLoc}
& \multicolumn{4}{c}{IPLoc-ID} \\
Backbone
& $N$=1 & $N$=2 & $N$=4 & $N$=8
& $N$=1 & $N$=2 & $N$=4 & $N$=8 \\
\hline
LLaVA1.5-7B
& 0.345 & 0.375 & 0.397 & \textbf{0.065}
& 0.348 & 0.379 & 0.369 & 0.064 \\

Gemma3-12B
& 0.377 & 0.395 & 0.406 & 0.444
& 0.382 & 0.442 & 0.422 & \textbf{0.450} \\

Qwen2-VL-7B
& 0.501 & 0.536 & 0.561 & \textbf{0.580}
& 0.503 & 0.535 & 0.571 & 0.580 \\

Qwen3-VL-8B
& 0.632 & 0.675 & 0.694 & 0.711
& 0.637 & 0.673 & 0.698 & \textbf{0.714} \\

Qwen3-VL-32B
& 0.650 & 0.702 & 0.716 & 0.728
& 0.639 & 0.701 & 0.723 & \textbf{0.729} \\

Qwen3-VL-235B
& 0.646 & 0.691 & 0.704 & 0.742
& 0.652 & 0.686 & 0.718 & \textbf{0.753} \\
\Xhline{1.2pt}
\end{tabular}
\\ \vspace{6pt}
%
\begin{tabular}{l|P{\pw}P{\pw}P{\pw}P{\pw}|P{\pw}P{\pw}P{\pw}P{\pw}}
\Xhline{1.2pt}
\multicolumn{1}{c|}{}
& \multicolumn{8}{c}{F1-score ($\uparrow$)} \\
\cline{2-9}
& \multicolumn{4}{c|}{IPLoc}
& \multicolumn{4}{c}{IPLoc-ID} \\
Backbone
& $N$=1 & $N$=2 & $N$=4 & $N$=8
& $N$=1 & $N$=2 & $N$=4 & $N$=8 \\
\hline
LLaVA1.5-7B
& 0.664 & \textbf{0.667} & \textbf{0.667} & 0.650
& 0.570 & 0.453 & 0.523 & 0.611 \\

Gemma3-12B
& 0.667 & 0.666 & 0.666 & 0.666
& 0.929 & 0.939 & 0.920 & \textbf{0.946} \\

Qwen2-VL-7B
& 0.667 & 0.666 & 0.667 & 0.667
& 0.943 & 0.963 & 0.973 & \textbf{0.985} \\

Qwen3-VL-8B
& 0.667 & 0.667 & 0.667 & 0.667
& 0.924 & 0.973 & 0.982 & \textbf{0.993} \\

Qwen3-VL-32B
& 0.668 & 0.667 & 0.667 & 0.667
& 0.950 & 0.968 & 0.985 & \textbf{0.996} \\

Qwen3-VL-235B
& 0.665 & 0.667 & 0.667 & 0.667
& 0.956 & 0.967 & 0.982 & \textbf{0.986} \\
\Xhline{1.2pt}
\end{tabular}
%
%
\end{table}
%
%
%

\begin{figure}
\centering
\small
\begin{tabular}{ccc}
\includegraphics[width=0.31\linewidth]{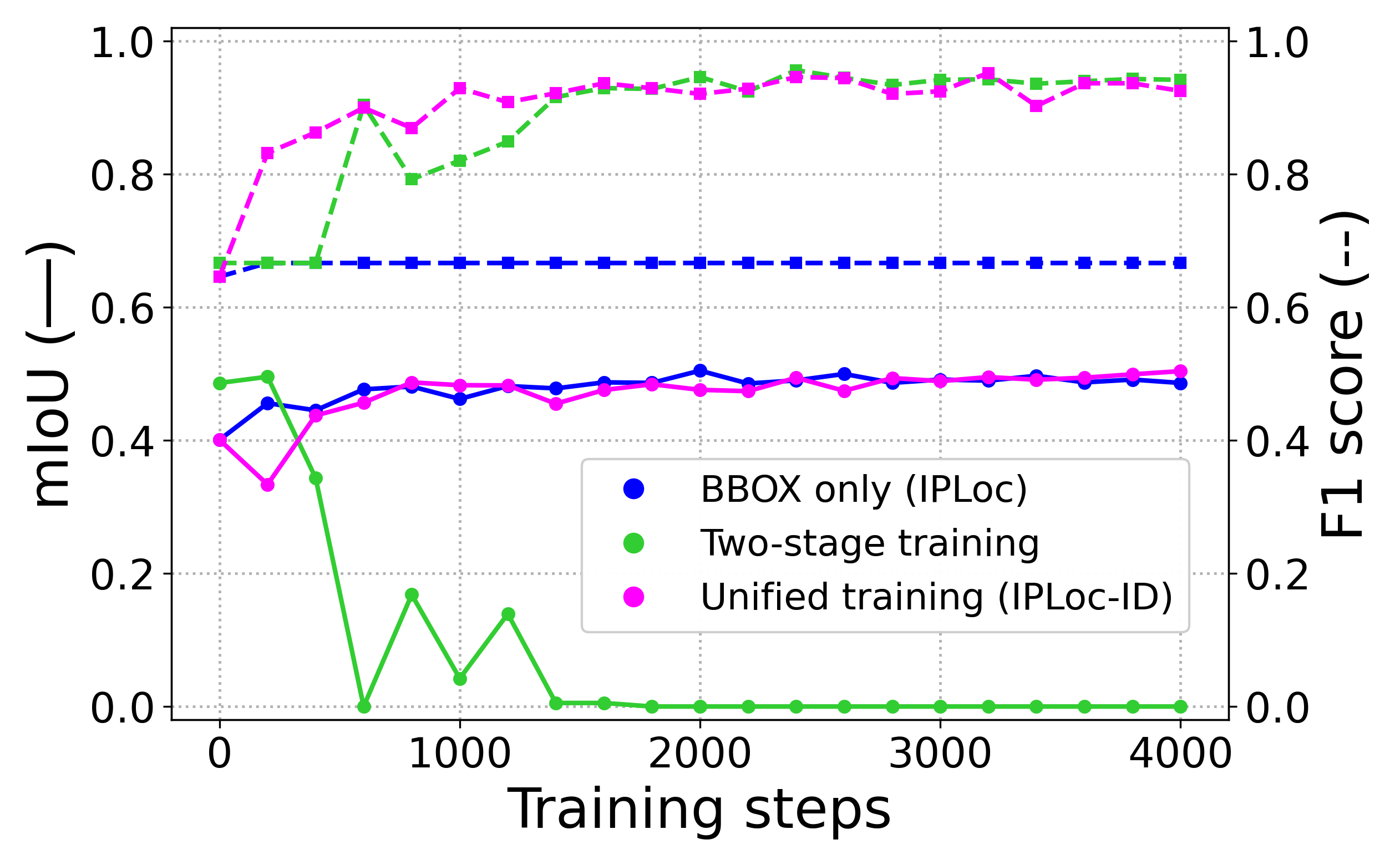} &
\includegraphics[width=0.31\linewidth]{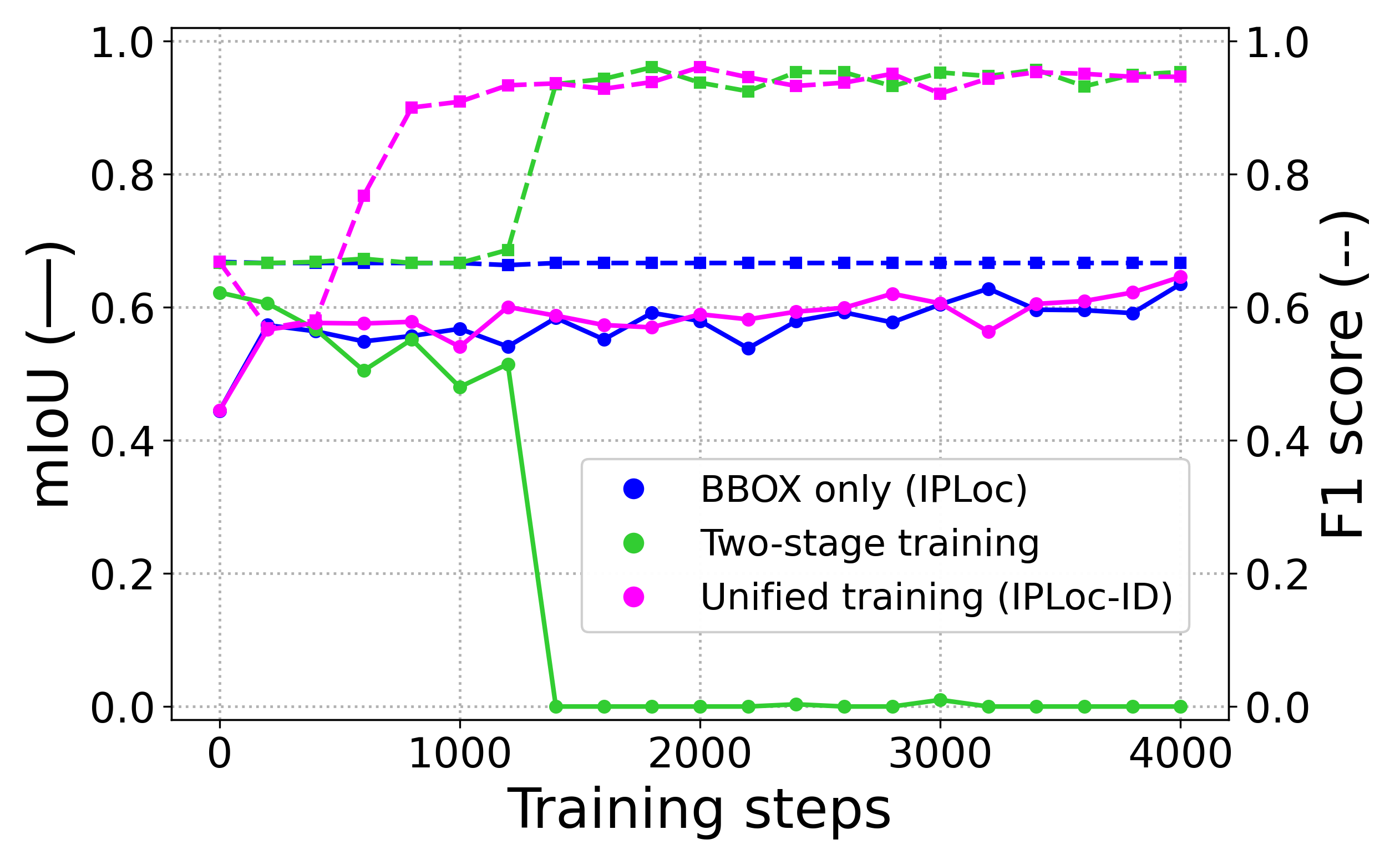} &
\includegraphics[width=0.31\linewidth]{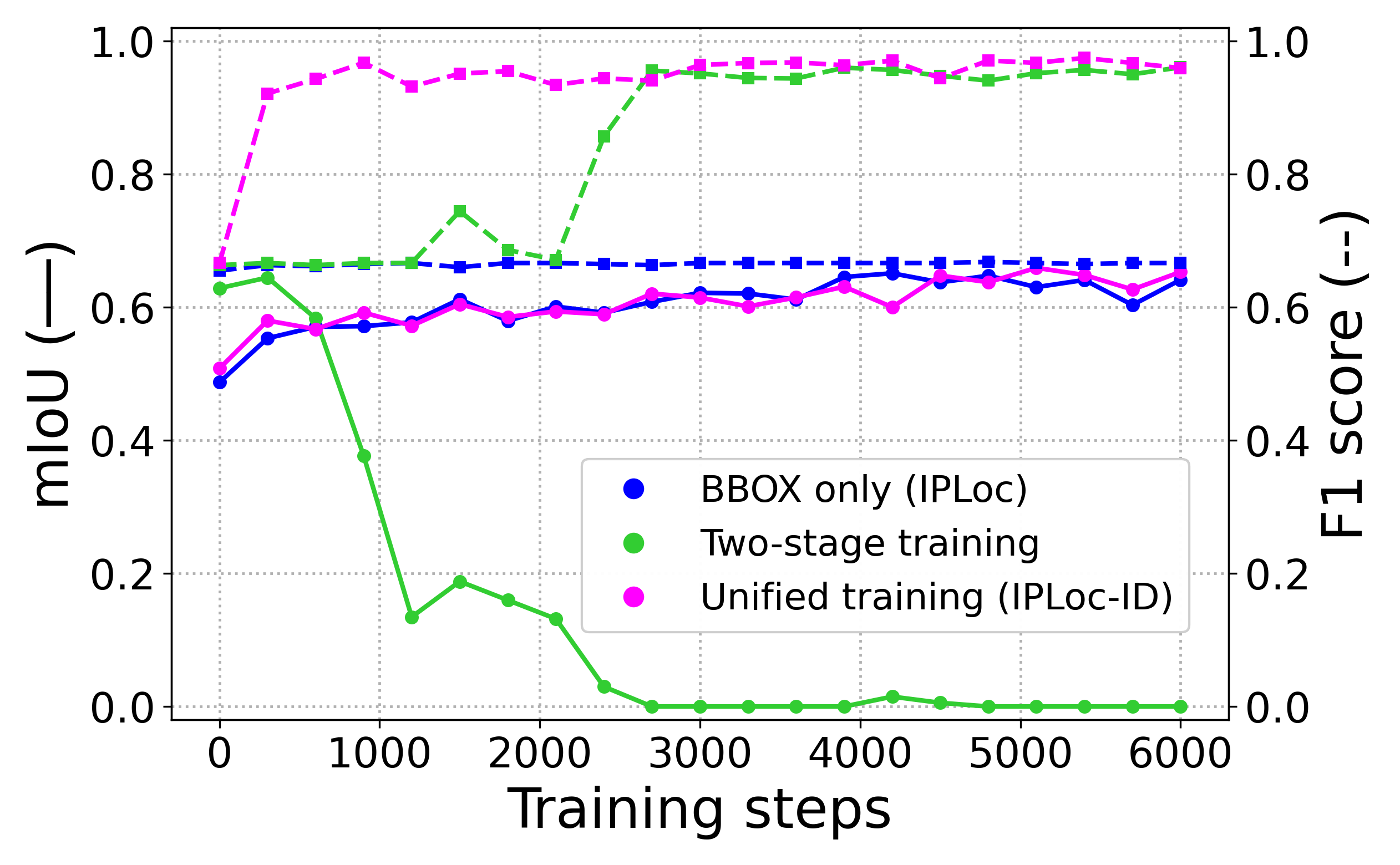} \\
(1) Qwen2-VL-7B & (2) Qwen3-VL-8B & (3) Qwen3-VL-32B
\end{tabular}
\caption{\textbf{[Training curves]} The mIoU (solid line) and F1-score (dotted line) curves for the LaSOT test set during training based on different backbones trained using (blue) only BBOX loss (IPLoc), (green) two-stage training, and (magenta) the proposed unified loss.}
\label{fig:unified-loss_vs_two-stage}
\end{figure}
%
%
%

%
%
%
\subsection{Ablation Studies} \label{sec:results:ablation}
\subsubsection{Unified objective vs.\ two-stage training}
The proposed IPLoc-ID uses the unified objective in Eq.~(\ref{eq:iplocid:objective-f}) to generate the complete output sequence in Eq.~(\ref{eq:iplocid:generated-y}), 
$\iplocidoutput=\langle \iplocidbbox \rangle \langle \iplocidquery \rangle \langle \iplocidanswer \rangle$. 
A natural question is whether this framework is merely an additional identification stage built upon a pretrained IPLoc model.
To answer this question, we compare IPLoc, IPLoc-ID, and a two-stage training strategy in Figure~\ref{fig:unified-loss_vs_two-stage}.
\textbf{Two-stage training} first trains the model for BBOX localization and then re-trains it only for identification, with the second-stage target sequence
$\langle \iplocidquery \rangle \langle \iplocidtargetanswer(x) \rangle$.
\par
Figure~\ref{fig:unified-loss_vs_two-stage} shows the mIoU and F1-score curves during training.
For the Two-stage model, only the second-stage training process starting from the pretrained IPLoc model is shown.
IPLoc improves mIoU but not F1-score, because it has no identification objective.
The Two-stage model improves F1-score, but its mIoU rapidly decreases, indicating catastrophic forgetting~\cite{mccloskey1989catastrophic,kirkpatrick2017overcoming} of localization ability.
In contrast, IPLoc-ID improves both mIoU and F1-score simultaneously.
These results show that IPLoc-ID is not merely an additional identification stage, but a unified framework for jointly learning localization and identification.
%

\def \pw {32pt}
\def \qw {38pt}
\begin{table}[htb]
\centering
\caption{\textbf{[Unified objective vs. conditional branching]} mIoU and F1-score on the 1-shot LaSOT test set for backbones under different training frameworks.}
\label{tab:unified_vs_branch}
\scriptsize
\begin{tabular}{l | P{\pw}P{\qw}P{\pw} | P{\pw}P{\qw}P{\pw}}
\Xhline{1.2pt}
& \multicolumn{3}{c|}{mIoU ($\uparrow$)} & \multicolumn{3}{c}{F1-score ($\uparrow$)} \\
Backbone & IPLoc & IPLoc-ID & Branch & IPLoc & IPLoc-ID & Branch \\
\hline
Qwen2-VL-7B  & 0.501 & \textbf{0.503} & 0.446 & 0.667 & \textbf{0.943} & 0.937 \\
Qwen3-VL-8B  & 0.632 & \textbf{0.637} & 0.595 & 0.667 & 0.924 & \textbf{0.955} \\
Qwen3-VL-32B & \textbf{0.644} & 0.639 & 0.620 & 0.670 & 0.950 & \textbf{0.952} \\
\Xhline{1.2pt}
\end{tabular}
%
%
\end{table}

%
%
%
\subsubsection{Unified objective vs.\ conditional branching}
We compare IPLoc-ID with another training strategy that directly generates different responses for positive and negative query images.
Given the same input sequence $x$, this baseline learns a conditionally branched response:
%
\begin{equation} \label{eq:conditional-branching}
y_{\mathrm{branch}} =
\begin{cases}
\langle \iplocidbbox \rangle, & \text{if } \delta(x)=1, \\
\langle \iplocidanswer \rangle, & \text{if } \delta(x)=0,
\end{cases}
\end{equation}
%
where $\delta(x)=1$ indicates that the query image contains the reference object instance, and $\delta(x)=0$ otherwise.
In our implementation, we use $\langle \iplocidanswer \rangle=\texttt{``Not found.''}$ as the negative response and refer to this variant as the \textbf{Conditional Branching} baseline.
\par
This formulation collapses bounding-box prediction and identification into a single conditional generation step.
Although straightforward, it requires the model to decide whether to localize or reject the query image before generating the output.
Moreover, generating a fixed negative response is easier than generating continuous-valued bounding-box coordinates, which tends to bias the model toward negative responses.
As shown in Table~\ref{tab:unified_vs_branch}, this results in degraded mIoU despite high F1-scores.
In contrast, IPLoc-ID first generates a reference-conditioned candidate BBOX, then performs identification through the self-posed query.
This sequential decomposition enables identification without sacrificing box localization accuracy, as shown in Table~\ref{tab:unified_vs_branch}.
%
%
%

\begin{table}[t]
\centering
\caption{\textbf{[Self-posed queries used in ablation]} Query texts used for the self-posed query ablation study.}
\label{tab:ablation:self-posed-query:texts}
\vspace{8pt}
\scriptsize
\begin{tabular}{p{0.22\linewidth}|p{0.69\linewidth}}
\Xhline{1.2pt}
Self-posed query & Prompt texts \\
\hline
Query \#1 & \texttt{Do all these boxes have the same object?} \\ 
Query \#2 & \texttt{Do all these boxes contain the same object instance?} \\ 
Query \#3 & \texttt{Is there a single shared object across all these boxes?} \\ 
Query \#4 & \texttt{Do all these boxes enclose the same object?} \\
\Xhline{1.2pt}
\end{tabular}
\vspace{-12pt} 
\end{table}

\def \pw {17pt}
\begin{table}[t]
\centering
\caption{\textbf{[Ablation on self-posed query]} mIoU and F1-score on the LaSOT test set for IPLoc-ID with different self-posed queries.}
\label{tab:ablation:self-posed-query:metrics}
%
%
\scriptsize
\begin{tabular}{l | P{\pw}P{\pw}P{\pw}P{\pw} | P{\pw}P{\pw}P{\pw}P{\pw}}
\Xhline{1.2pt}
& \multicolumn{4}{c|}{mIoU ($\uparrow$)} & \multicolumn{4}{c}{F1-score ($\uparrow$)} \\
Self-posed query
& \#1 & \#2 & \#3 & \#4
& \#1 & \#2 & \#3 & \#4 \\
\hline
Qwen2-VL-7B + IPLoc-ID
& \textbf{0.503} & 0.489 & 0.489 & 0.490
& 0.943 & \textbf{0.944} & 0.944 & 0.940 \\
Qwen3-VL-8B + IPLoc-ID
& 0.637 & 0.623 & 0.623 & \textbf{0.639}
& 0.924 & 0.898 & \textbf{0.944} & 0.942 \\
Qwen3-VL-32B + IPLoc-ID
& 0.639 & \textbf{0.668} & 0.657 & 0.645
& 0.950 & \textbf{0.966} & 0.965 & 0.952 \\
\Xhline{1.2pt}
\end{tabular}
\vspace{-12pt} 
\end{table}

%
%
%
\subsubsection{Ablation on self-posed query} \label{sec:results:ablation:self-posed-query}
We ablate the concrete wording of the self-posed query introduced in Section~\ref{sec:method:iplocid}.
The four query texts compared in this experiment are listed in Table~\ref{tab:ablation:self-posed-query:texts}.
They cover generic object consistency, instance-level identity, global set consistency, and spatial enclosure.
\par
Table~\ref{tab:ablation:self-posed-query:metrics} reports the mIoU and F1-score obtained with these four self-posed queries.
The results show that IPLoc-ID is robust to the specific wording of $\langle \iplocidquery \rangle$.
This suggests that, during fine-tuning, the model learns both to generate the self-posed query and to use it as a cue for the subsequent answer component $\langle \iplocidanswer \rangle$.
In this paper, we use Query \#1 as the recommended implementation because it is the most basic formulation and shows stable performance.
\subsubsection{Ablation on loss terms with label noise}
The proposed IPLoc-ID consists of three components: the box localization term, pseudo-label-based label noise inherited from IPLoc, and the identification term.
Table~\ref{tab:ablation-label-noise} analyzes different combinations of these components.
The first and second columns correspond to IPLoc without and with label noise, respectively.
The third column removes label noise from IPLoc-ID, and the fourth column shows the full IPLoc-ID.
%
The results reconfirm that label noise improves mIoU not only for Qwen2 but also for the Qwen3 series, and we therefore use pseudo-labeling in our final models.
More importantly, the identification term consistently improves F1-score, regardless of the use of label noise.
%
%

%
%
\def\pw{23pt}
\def\qw{56pt}
\begin{table}[t]
\centering
\caption{\textbf{[Ablation on loss terms with label noise]} mIoU and F1-score on the LaSOT test set for different backbones. The four columns in each metric block correspond to different combinations of loss components.}
\label{tab:ablation-label-noise}
\scriptsize
\begin{tabular}{Q{\qw}|*{4}{P{\pw}}|*{4}{P{\pw}}}
\Xhline{1.2pt}
& \multicolumn{4}{c|}{mIoU ($\uparrow$)} & \multicolumn{4}{c}{F1-score ($\uparrow$)} \\
\hline
box localization & 
$\checkmark$ & $\checkmark$ & $\checkmark$ & $\checkmark$ & 
$\checkmark$ & $\checkmark$ & $\checkmark$ & $\checkmark$ \\
+ label noise   &   & $\checkmark$ &   & $\checkmark$ &   & $\checkmark$ &   & $\checkmark$ \\
+ identification &   &   & $\checkmark$ & $\checkmark$ &   &   & $\checkmark$ & $\checkmark$ \\
\hline
Qwen2-VL-7B  & 0.495 & 0.501 & 0.491 & 0.503 & 0.666 & 0.667 & 0.939 & 0.943 \\
Qwen3-VL-8B  & 0.595 & 0.632 & 0.601 & 0.637 & 0.667 & 0.667 & 0.933 & 0.924 \\
Qwen3-VL-32B & 0.631 & 0.650 & 0.637 & 0.639 & 0.667 & 0.668 & 0.964 & 0.950 \\
\Xhline{1.2pt}
\end{tabular}
\renewcommand{\arraystretch}{0.8}
\begin{tabular}{Q{\qw}*{4}{P{\pw}}*{4}{P{\pw}}}
& & \scriptsize (IPLoc)  & & \scriptsize (Ours)&  & \scriptsize (IPLoc) & & \scriptsize (Ours)  \\
\end{tabular}
\vspace{-12pt} 
\end{table}
%
%
%

%
%
%
\subsection{Comprehensive Comparison with State-of-the-Art Methods} \label{sec:results:final}
We now present the main experimental results on LaSOT, PDM, GOT-10K, and VastTrack in Tables~\ref{tab:main-results-lasot}--\ref{tab:main-results-vasttrack}.
In these tables, the best value for each $N$-shot setting within each block is highlighted in bold, while the best value within each block is underlined.
We compare three variants for each backbone: the VLM with instruction prompting, IPLoc, and IPLoc-ID.
For LLaVA1.5-7B and Gemma3-12B, the instruction-prompting baseline is denoted by adding ``+ prompt''.
For brevity, we omit ``+ prompt'' for the Qwen series and denote the prompted backbone simply by the model name.
For Qwen3-VL-235B, we report a single independent trial due to its high computational cost.
For externally prompted VLM baselines, we use a structured two-step instruction prompt selected from a preliminary prompt pretest.
The details of this pretest are provided in \ref{sec:appendix-prompt-pretest}.
\par
We also include the following related algorithms.
\textbf{Grounding-DINO}~\cite{liu2024groundingdino} is used as a general-purpose object detector.
\textbf{Florence-2}~\cite{xiao2024florence} is used as a multi-task VLM with OD prompting.
Both use only the target image and label, so we report them only under the 1-shot setting.
\textbf{VFM}~\cite{feng2026fsodvfm} is used as a state-of-the-art FSOD baseline with all $N$-shot reference data as support data.
\textbf{No-Time-To-Train}~\cite{espinosa2025notime} constructs a memory bank from the reference data for each test case and applies it to the target image.
\textbf{LLaVA1.5-7B}~\cite{liu2023llava} with instruction prompting is included as an early conversational VLM baseline.
\textbf{IPLoc 7B (official)}~\cite{doveh2025iploc} is the official Qwen2-VL-7B-based IPLoc model.
\par 
The results on the LaSOT test set in Table~\ref{tab:main-results-lasot} evaluate generalization to unseen classes within the same domain, as described in Section~\ref{sec:method:dataset}. 
In this setting, IPLoc-ID maintains BBOX localization accuracy comparable to IPLoc in terms of mIoU, while substantially improving F1-score.
Recall that an F1-score of approximately $0.667$ corresponds to an all-positive response under a balanced positive/negative test set.
Thus, the F1-scores of IPLoc and several localization-only baselines indicate limited instance-level identification capability.
In contrast, IPLoc-ID consistently achieves high F1-scores, demonstrating that it effectively rejects negative query images.
\par 
The results on PDM and GOT-10K in Tables~\ref{tab:main-results-pdm} and~\ref{tab:main-results-got10k} evaluate generalization to unseen domains.
In terms of mIoU, IPLoc and IPLoc-ID achieve satisfactory localization performance, although No-Time-To-Train shows particularly strong mIoU on PDM.
This indicates that IPLoc-based methods can benefit from the general visual reasoning ability of large-scale pretrained VLMs.
For F1-score, VFM achieves high values on GOT-10K, likely because the negative examples are out-of-class images, making the task closer to category-level FSOD.
The F1-scores of IPLoc remain close to the all-positive baseline, indicating that IPLoc tends to return positive detections even for out-of-class negative examples.
In contrast, IPLoc-ID achieves the highest F1-scores on these two sets, demonstrating strong identification ability under unseen-domain settings.
\par 
The VastTrack results in Table~\ref{tab:main-results-vasttrack} are particularly important because the negative examples are in-class distractors.
This setting requires instance-level identification and evaluates unseen-domain generalization without fine-tuning on that domain.
Even under this challenging setting, IPLoc-ID shows consistent improvements in F1-score while maintaining competitive mIoU.
\par 
Regarding the reference algorithms, LLaVA1.5-7B with instruction prompting shows substantially lower mIoU than the other algorithms.
This is because early VLMs such as LLaVA1.5-7B often ignore localization-oriented instructions and generate image captions or free-form descriptions, as also reported in the original IPLoc study~\cite{doveh2025iploc}.
The official IPLoc 7B model shows behavior similar to our reproduced Qwen2-VL-7B + IPLoc model across the tested datasets.
This illustrates that the original IPLoc tends to produce false-positive detections on negative query images and supports the validity of our reproduced IPLoc implementation.
VLMs with instruction prompting can solve the POIL task to some extent, especially with stronger backbone models.
However, IPLoc-ID achieves better overall performance in both mIoU and F1-score.
\par
Importantly, these results across datasets and backbone models empirically support the formulations in Section~\ref{sec:method}. 
The localization-only IPLoc variants remain close to the all-positive F1-score baseline, consistent with Eq.~(\ref{eq:iploc:f-for-negative}): IPLoc tends to return an element of $\mathcal{B}$ even for negative query images, whereas the ideal POIL mapping in Eq.~(\ref{eq:oil-task-fx}) requires rejection. 
In contrast, IPLoc-ID substantially improves F1-score while preserving mIoU, showing that the interpreted prediction in Eq.~(\ref{eq:iplocid:interpreted}) better approximates the ideal mapping in Eq.~(\ref{eq:oil-task-fx}). 
Thus, the results support the intended transition from localization-only prediction to identification-aware personalized object localization.

\clearpage

%
%
%
\def \pw {18pt}
\def \qw {100pt}
\begin{table}[htb]
\centering
\caption{\textbf{[Quantitative comparison on the LaSOT test set]} mIoU and F1-score under different $N$-shot settings for various algorithms.}
\label{tab:main-results-lasot}
\scriptsize
\begin{tabular}{p{\qw} | P{\pw}P{\pw}P{\pw}P{\pw} | P{\pw}P{\pw}P{\pw}P{\pw}}
\Xhline{1.2pt}
& \multicolumn{4}{c|}{mIoU ($\uparrow$)} & \multicolumn{4}{c}{F1-score ($\uparrow$)} \\
Algorithms  & $N$=1 & $N$=2 & $N$=4 & $N$=8 & $N$=1 & $N$=2 & $N$=4 & $N$=8  \\
\hline
(OD) Grounding-DINO      & 0.222 & - & - & - & 0.416 & - & - & - \\
(OD) Florence-2          & 0.306 & - & - & - & 0.511 & - & - & - \\
(FSOD) VFM               & 0.605 & 0.593 & 0.617 & 0.601 & 0.656 & \underline{\textbf{0.665}} & 0.662 & \underline{\textbf{0.665}} \\
(FSOD) No-Time-To-Train  & \underline{\textbf{0.644}} & \textbf{0.627} & \textbf{0.618} & \textbf{0.617} & \textbf{0.663} & 0.657 & 0.660 & 0.657 \\
LLaVA1.5-7B + prompt     & 0.005 & 0.159 & 0.111 & 0.000 & 0.065 & 0.624 & 0.524 & 0.014 \\
IPLoc 7B (official)      & 0.515 & 0.529 & 0.541 & 0.527 & 0.652 & 0.663 & \textbf{0.663} & 0.659 \\
\hline
Gemma3-12B (+ prompt)    & 0.139 & 0.185 & 0.209 & 0.140 & 0.869 & 0.843 & 0.795 & 0.535 \\
Gemma3-12B + IPLoc       & 0.377 & 0.395 & 0.406 & 0.444 & 0.667 & 0.666 & 0.666 & 0.666 \\
Gemma3-12B + IPLoc-ID    & \textbf{0.382} & \textbf{0.442} & \textbf{0.422} & \underline{\textbf{0.450}} & \textbf{0.929} & \textbf{0.939} & \textbf{0.920} & \underline{\textbf{0.946}} \\
\hline
Qwen2-VL-7B              & 0.247 & 0.306 & 0.297 & 0.319 & 0.584 & 0.489 & 0.426 & 0.488 \\
Qwen2-VL-7B + IPLoc      & 0.501 & \textbf{0.536} & 0.561 & \underline{\textbf{0.580}} & 0.667 & 0.666 & 0.667 & 0.667 \\
Qwen2-VL-7B + IPLoc-ID   & \textbf{0.503} & 0.535 & \textbf{0.571} & 0.580 & \textbf{0.943} & \textbf{0.963} & \textbf{0.973} & \underline{\textbf{0.985}} \\
\hline
Qwen3-VL-8B              & 0.511 & 0.558 & 0.552 & 0.559 & 0.746 & 0.779 & 0.810 & 0.771 \\
Qwen3-VL-8B + IPLoc      & 0.632 & \textbf{0.675} & 0.694 & 0.711 & 0.667 & 0.667 & 0.667 & 0.667 \\
Qwen3-VL-8B + IPLoc-ID   & \textbf{0.637} & 0.673 & \textbf{0.698} & \underline{\textbf{0.714}} & \textbf{0.924} & \textbf{0.973} & \textbf{0.982} & \underline{\textbf{0.993}} \\
\hline
Qwen3-VL-32B             & 0.541 & 0.561 & 0.572 & 0.585 & 0.835 & 0.889 & 0.883 & 0.927 \\
Qwen3-VL-32B + IPLoc     & \textbf{0.650} & \textbf{0.702} & 0.716 & 0.728 & 0.668 & 0.667 & 0.667 & 0.667 \\
Qwen3-VL-32B + IPLoc-ID  & 0.639 & 0.701 & \textbf{0.723} & \underline{\textbf{0.729}} & \textbf{0.950} & \textbf{0.968} & \textbf{0.985} & \underline{\textbf{0.996}} \\
\hline
Qwen3-VL-235B            & 0.430 & 0.561 & 0.584 & 0.588 & 0.866 & 0.840 & 0.884 & 0.935 \\
Qwen3-VL-235B + IPLoc    & 0.646 & \textbf{0.691} & 0.704 & 0.742 & 0.665 & 0.667 & 0.667 & 0.667 \\
Qwen3-VL-235B + IPLoc-ID & \textbf{0.652} & 0.686 & \textbf{0.718} & \underline{\textbf{0.753}} & \textbf{0.956} & \textbf{0.967} & \textbf{0.982} & \underline{\textbf{0.986}} \\
\Xhline{1.2pt}
\end{tabular}
\vspace{-18pt} 
\end{table}
%
%
%

%
%
\def \pw {47pt}
\def \qw {100pt}
\begin{table}[htb]
\centering
\caption{\textbf{[Quantitative comparison on PDM]} mIoU and F1-score under different $N$-shot settings for various algorithms.}
\label{tab:main-results-pdm}
\scriptsize
\begin{tabular}{p{\qw} | P{\pw}P{\pw} | P{\pw}P{\pw}}
\Xhline{1.2pt}
& \multicolumn{2}{c|}{mIoU ($\uparrow$)} & \multicolumn{2}{c}{F1-score ($\uparrow$)} \\
Algorithms  & $N$=1 & $N$=2 & $N$=1 & $N$=2  \\
\hline
(OD) Grounding-DINO      & 0.213 & - & 0.498 & - \\
(OD) Florence-2          & 0.174 & - & 0.467 & - \\
(FSOD) VFM               & 0.264 & 0.287 & 0.658 & 0.681 \\
(FSOD) No-Time-To-Train  & \textbf{0.556} & \underline{\textbf{0.621}} & 0.636 & 0.655 \\
LLaVA1.5-7B + prompt     & 0.009 & 0.122 & 0.100 & 0.644 \\
IPLoc 7B (official)      & 0.315 & 0.343 & \underline{\textbf{0.722}} & \textbf{0.713} \\
\hline
Gemma3-12B (+ prompt)    & 0.117 & 0.142 & 0.760 & 0.663 \\
Gemma3-12B + IPLoc       & 0.134 & 0.177 & 0.667 & 0.666 \\
Gemma3-12B + IPLoc-ID    & \textbf{0.134} & \underline{\textbf{0.201}} & \textbf{0.932} & \underline{\textbf{0.959}} \\
\hline
Qwen2-VL-7B              & 0.204 & 0.229 & 0.380 & 0.455 \\
Qwen2-VL-7B + IPLoc      & \textbf{0.318} & \underline{\textbf{0.367}} & 0.666 & 0.667 \\
Qwen2-VL-7B + IPLoc-ID   & 0.316 & 0.351 & \textbf{0.976} & \underline{\textbf{0.988}} \\
\hline
Qwen3-VL-8B              & 0.363 & 0.398 & 0.861 & 0.914 \\
Qwen3-VL-8B + IPLoc      & 0.436 & \underline{\textbf{0.487}} & 0.667 & 0.667 \\
Qwen3-VL-8B + IPLoc-ID   & \textbf{0.439} & \underline{\textbf{0.487}} & \textbf{0.941} & \underline{\textbf{0.987}} \\
\hline
Qwen3-VL-32B             & 0.391 & 0.440 & 0.699 & 0.654 \\
Qwen3-VL-32B + IPLoc     & 0.437 & 0.473 & 0.667 & 0.667 \\
Qwen3-VL-32B + IPLoc-ID  & \textbf{0.455} & \underline{\textbf{0.507}} & \textbf{0.986} & \underline{\textbf{0.995}} \\
\hline
Qwen3-VL-235B            & 0.368 & 0.387 & 0.778 & 0.633 \\
Qwen3-VL-235B + IPLoc    & \textbf{0.445} & 0.481 & 0.668 & 0.666 \\
Qwen3-VL-235B + IPLoc-ID & \textbf{0.445} & \underline{\textbf{0.536}} & \textbf{0.970} & \underline{\textbf{0.976}} \\
\Xhline{1.2pt}
\end{tabular}
\vspace{-18pt} 
\end{table}

\clearpage

%
%
%
\def \pw {18pt}
\def \qw {100pt}
\begin{table}[htb]
\centering
\caption{\textbf{[Quantitative comparison on GOT-10K]} mIoU and F1-score under different $N$-shot settings for various algorithms.}
\label{tab:main-results-got10k}
\scriptsize
\begin{tabular}{p{\qw} | P{\pw}P{\pw}P{\pw}P{\pw} | P{\pw}P{\pw}P{\pw}P{\pw}}
\Xhline{1.2pt}
& \multicolumn{4}{c|}{mIoU ($\uparrow$)} & \multicolumn{4}{c}{F1-score ($\uparrow$)} \\
Algorithms  & $N$=1 & $N$=2 & $N$=4 & $N$=8 & $N$=1 & $N$=2 & $N$=4 & $N$=8  \\
\hline
(OD) Grounding-DINO      & 0.012 & - & - & - & 0.054 & - & - & - \\
(OD) Florence-2          & 0.039 & - & - & - & 0.115 & - & - & - \\
(FSOD) VFM               & 0.680 & 0.717 & 0.736 & 0.764 & \textbf{0.942} & \textbf{0.946} & \textbf{0.946} & \underline{\textbf{0.960}} \\
(FSOD) No-Time-To-Train  & \textbf{0.762} & \textbf{0.768} & \underline{\textbf{0.769}} & \underline{\textbf{0.769}} & 0.672 & 0.670 & 0.672 & 0.670 \\
LLaVA1.5-7B + prompt     & 0.000 & 0.139 & 0.082 & 0.000 & 0.000 & 0.529 & 0.410 & 0.043 \\
IPLoc 7B (official)      & 0.481 & 0.501 & 0.516 & 0.587 & 0.675 & 0.677 & 0.668 & 0.668 \\
\hline
Gemma3-12B (+ prompt)    & 0.201 & 0.198 & 0.234 & 0.295 & \textbf{0.966} & 0.922 & 0.899 & 0.751 \\
Gemma3-12B + IPLoc       & \textbf{0.504} & 0.485 & \textbf{0.511} & 0.572 & 0.667 & 0.667 & 0.666 & 0.667 \\
Gemma3-12B + IPLoc-ID    & \textbf{0.504} & \textbf{0.534} & 0.504 & \underline{\textbf{0.607}} & 0.942 & \textbf{0.940} & \textbf{0.910} & \underline{\textbf{0.972}} \\
\hline
Qwen2-VL-7B              & 0.231 & 0.235 & 0.274 & 0.434 & 0.731 & 0.602 & 0.414 & 0.669 \\
Qwen2-VL-7B + IPLoc      & \textbf{0.497} & 0.527 & 0.558 & 0.634 & 0.667 & 0.667 & 0.667 & 0.667 \\
Qwen2-VL-7B + IPLoc-ID   & 0.496 & \textbf{0.532} & \textbf{0.570} & \underline{\textbf{0.643}} & \textbf{0.970} & \textbf{0.949} & \textbf{0.944} & \underline{\textbf{0.997}} \\
\hline
Qwen3-VL-8B              & 0.667 & 0.690 & 0.649 & 0.676 & \textbf{0.967} & \textbf{0.983} & \textbf{0.989} & 0.985 \\
Qwen3-VL-8B + IPLoc      & \textbf{0.763} & 0.771 & \textbf{0.786} & \underline{\textbf{0.836}} & 0.667 & 0.667 & 0.667 & 0.667 \\
Qwen3-VL-8B + IPLoc-ID   & 0.747 & \textbf{0.772} & 0.785 & 0.828 & 0.943 & 0.967 & 0.973 & \underline{\textbf{0.997}} \\
\hline
Qwen3-VL-32B             & 0.686 & 0.693 & 0.671 & 0.668 & 0.947 & 0.949 & 0.956 & 0.983 \\
Qwen3-VL-32B + IPLoc     & \textbf{0.738} & \textbf{0.756} & \textbf{0.796} & 0.842 & 0.668 & 0.664 & 0.667 & 0.667 \\
Qwen3-VL-32B + IPLoc-ID  & \textbf{0.738} & 0.742 & 0.777 & \underline{\textbf{0.854}} & \textbf{0.993} & \textbf{0.986} & \textbf{0.982} & \underline{\textbf{0.997}} \\
\hline
Qwen3-VL-235B            & 0.616 & 0.671 & 0.625 & 0.737 & 0.951 & 0.916 & 0.915 & 0.929 \\
Qwen3-VL-235B + IPLoc    & 0.736 & 0.785 & \textbf{0.806} & 0.860 & 0.667 & 0.667 & 0.668 & 0.667 \\
Qwen3-VL-235B + IPLoc-ID & \textbf{0.764} & \textbf{0.795} & 0.795 & \underline{\textbf{0.869}} & \underline{\textbf{1.000}} & \textbf{0.974} & \textbf{0.935} & \underline{\textbf{1.000}} \\
\Xhline{1.2pt}
\end{tabular}
\vspace{-18pt} 
\end{table}
%
%
%

%
%
\def \pw {18pt}
\def \qw {100pt}
\begin{table}[htb]
\centering
\caption{\textbf{[Quantitative comparison on VastTrack]} mIoU and F1-score under different $N$-shot settings for various algorithms.}
\label{tab:main-results-vasttrack}
\scriptsize
\begin{tabular}{p{\qw} | P{\pw}P{\pw}P{\pw}P{\pw} | P{\pw}P{\pw}P{\pw}P{\pw}}
\Xhline{1.2pt}
& \multicolumn{4}{c|}{mIoU ($\uparrow$)} & \multicolumn{4}{c}{F1-score ($\uparrow$)} \\
Algorithms  & $N$=1 & $N$=2 & $N$=4 & $N$=8 & $N$=1 & $N$=2 & $N$=4 & $N$=8  \\
\hline
(OD) Grounding-DINO      & 0.003 & - & - & - & 0.030 & - & - & - \\
(OD) Florence-2          & 0.026 & - & - & - & 0.101 & - & - & - \\
(FSOD) VFM               & 0.400 & 0.413 & 0.440 & 0.473 & \textbf{0.690} & \textbf{0.694} & \textbf{0.712} & \underline{\textbf{0.723}} \\
(FSOD) No-Time-To-Train  & \textbf{0.500} & \textbf{0.512} & \textbf{0.524} & \underline{\textbf{0.530}} & 0.646 & 0.648 & 0.650 & 0.650 \\
LLaVA1.5-7B + prompt     & 0.000 & 0.108 & 0.100 & 0.000 & 0.010 & 0.554 & 0.474 & 0.005 \\
IPLoc 7B (official)      & 0.284 & 0.316 & 0.333 & 0.427 & 0.660 & 0.664 & 0.662 & 0.663 \\
\hline
Gemma3-12B (+ prompt)    & 0.154 & 0.161 & 0.216 & 0.289 & 0.784 & 0.785 & 0.755 & 0.629 \\
Gemma3-12B + IPLoc       & \textbf{0.227} & 0.243 & 0.287 & 0.413 & 0.666 & 0.666 & 0.667 & 0.667 \\
Gemma3-12B + IPLoc-ID    & 0.223 & \textbf{0.267} & \textbf{0.291} & \underline{\textbf{0.415}} & \textbf{0.884} & \textbf{0.907} & \textbf{0.888} & \underline{\textbf{0.913}} \\
\hline
Qwen2-VL-7B              & 0.180 & 0.192 & 0.234 & 0.380 & 0.565 & 0.509 & 0.416 & 0.500 \\
Qwen2-VL-7B + IPLoc      & \textbf{0.335} & \textbf{0.357} & \textbf{0.382} & \underline{\textbf{0.472}} & 0.667 & 0.667 & 0.667 & 0.667 \\
Qwen2-VL-7B + IPLoc-ID   & 0.329 & 0.349 & 0.378 & 0.471 & \textbf{0.906} & \textbf{0.937} & \textbf{0.953} & \underline{\textbf{0.975}} \\
\hline
Qwen3-VL-8B              & 0.359 & 0.378 & 0.387 & 0.465 & 0.706 & 0.745 & 0.761 & 0.741 \\
Qwen3-VL-8B + IPLoc      & \textbf{0.429} & \textbf{0.451} & 0.477 & \underline{\textbf{0.578}} & 0.667 & 0.667 & 0.667 & 0.667 \\
Qwen3-VL-8B + IPLoc-ID   & 0.424 & 0.447 & \textbf{0.482} & 0.574 & \textbf{0.884} & \textbf{0.951} & \textbf{0.959} & \underline{\textbf{0.965}} \\
\hline
Qwen3-VL-32B             & 0.188 & 0.413 & 0.382 & 0.465 & 0.379 & 0.838 & 0.869 & 0.918 \\
Qwen3-VL-32B + IPLoc     & \textbf{0.443} & \textbf{0.472} & 0.507 & 0.596 & 0.667 & 0.668 & 0.667 & 0.667 \\
Qwen3-VL-32B + IPLoc-ID  & 0.437 & 0.465 & \textbf{0.508} & \underline{\textbf{0.615}} & \textbf{0.928} & \textbf{0.958} & \textbf{0.951} & \underline{\textbf{0.973}} \\
\hline
Qwen3-VL-235B            & 0.326 & 0.358 & 0.378 & 0.502 & 0.802 & 0.783 & 0.725 & 0.726 \\
Qwen3-VL-235B + IPLoc    & 0.427 & 0.456 & 0.506 & 0.618 & 0.667 & 0.667 & 0.667 & 0.667 \\
Qwen3-VL-235B + IPLoc-ID & \textbf{0.431} & \textbf{0.462} & \textbf{0.510} & \underline{\textbf{0.621}} & \textbf{0.930} & \textbf{0.954} & \textbf{0.963} & \underline{\textbf{0.982}} \\
\Xhline{1.2pt}
\end{tabular}
\vspace{-18pt} 
\end{table}

\clearpage

\par
For qualitative comparison, Figure~\ref{fig:incontext-inference-for-POIL} visualizes representative examples on the LaSOT test set.
The examples show that conventional OD/FSOD methods and localization-only IPLoc tend to produce false-positive detections on negative query images, whereas IPLoc-ID suppresses such false positives while preserving correct localization on positive query images.
Additional qualitative comparisons between the baseline IPLoc and the proposed IPLoc-ID on all test datasets are provided in~\ref{sec:appendix-qualitative}.
%
%
%
%
%
%
%
%
%
%
%
%
%

%
%
%
%
\section{Conclusion} \label{sec:conclusion}
In this paper, we generalized personalized object localization (POL) to personalized object identification and localization (POIL).
Unlike localization-only settings, POIL requires localizing the reference-conditioned object instance when it appears in the query image and rejecting the image otherwise.
Thus, POIL combines reference-conditioned instance-level localization with negative-query rejection.
For this task, we constructed POIL datasets from four public sources, including both positive and negative query images.
\par
We proposed IPLoc-ID as an in-context algorithm for POIL.
IPLoc-ID treats the BBOX generated by IPLoc as a candidate and verifies whether it corresponds to the reference object instance through a self-posed query and identification answer.
This design connects the input context, BBOX candidate, self-posed query, and final identification response as a single autoregressive sequence.
Through experiments, we demonstrated that IPLoc-ID substantially suppresses false-positive detections on negative query images while maintaining the localization performance of IPLoc.
In particular, the results on in-class negative examples show that IPLoc-ID is more effective for instance-level identification than conventional OD, FSOD, and localization-only IPLoc.
\par
The limitations of this study are as follows.
First, similar to IPLoc, our setting focuses on a single object in each query image and does not address simultaneous localization and identification of multiple target objects.
Second, inference is performed on individual images, and temporal consistency in videos or image sequences is not explicitly used.
Thus, the current framework does not fully exploit temporal information and motion consistency, which are important for applications such as video grounding and object tracking.
\par
For the broader research community, we expect the POIL formulation, customized datasets, evaluation protocol, and baseline comparisons introduced in this study to provide a useful foundation for future research on various personalized vision tasks, including personalized recognition, retrieval, grounding, and tracking under realistic query settings.
\par
As VLMs continue to advance rapidly, future work will extend IPLoc-ID to multi-object and video-level POIL tasks by further exploiting the multi-object localization and temporal reasoning abilities of increasingly capable VLMs through the proposed reference-conditioned identification framework.
\section*{Acknowledgment}
This work was supported by the Korea government (MSIT): IITP-RS-2021-II211341, Artificial Intelligence Graduate School, Chung-Ang University and NRF-RS-2025-25462275.

\section*{Data and code availability}
This study uses publicly available source datasets, including LaSOT, PDM/BURST, GOT-10K, and VastTrack.
We provide the inference code, dataset construction scripts, and minimal trained models at \url{https://github.com/kensuke-nakamura/iplocid}.
The training code and additional trained models will be made publicly available upon acceptance.

\bibliographystyle{elsarticle-num} 
\bibliography{mybib}

\clearpage

%
%
%
%
%
%
%
\setcounter{figure}{0}
\setcounter{table}{0}
\appendix
\section{Additional Experimental Results} \label{sec:appendix-additional-results}
\subsection{Pretest on instruction prompts} \label{sec:appendix-prompt-pretest}
The proposed IPLoc-ID is designed for in-context inference without updating the model on test data.
As an auxiliary analysis, we conducted a pretest on instruction prompt design for externally prompted VLMs.
We compared four prompts, whose concrete texts are shown in Figure~\ref{fig:ablation:instruction-prompts}.
Prompt \#1 is a two-line explicit format that specifies both the bounding box and identity response; Prompt \#2 is a compact one-line format of the form [x1,y1,x2,y2], YES/NO; Prompt \#3 is a structured two-step reasoning format that guides localization and identity verification; and Prompt \#4 is a minimal-constraint format.
Table~\ref{tab:prompt-pretest} reports the results on the 1-shot LaSOT test set.
Based on these results, we adopt Prompt \#3 for externally prompted VLMs in the comprehensive comparison.
%
%
%

%
%
\def \pw {18pt}
\begin{table}[htb]
\centering
\caption{\textbf{[Pretest on instruction prompts]} mIoU and F1-score on the 1-shot LaSOT test set for VLMs with different instruction prompts (\#1--\#4).}
\label{tab:prompt-pretest}
\scriptsize
\begin{tabular}{l | P{\pw}P{\pw}P{\pw}P{\pw} | P{\pw}P{\pw}P{\pw}P{\pw}}
\Xhline{1.2pt}
& \multicolumn{4}{c|}{mIoU ($\uparrow$)}
& \multicolumn{4}{c}{F1-score ($\uparrow$)} \\
Instruction prompt & \#1 & \#2 & \#3 & \#4 & \#1 & \#2 & \#3 & \#4 \\
\hline
Gemma-3-12B + prompt 
& 0.146 & 0.135 & 0.139 & \textbf{0.154} 
& 0.675 & 0.661 & \textbf{0.869} & 0.667 \\
Qwen2-VL-7B + prompt 
& 0.226 & 0.167 & \textbf{0.247} & 0.211 
& 0.650 & 0.655 & 0.584 & \textbf{0.667} \\
Qwen3-VL-8B + prompt 
& 0.503 & 0.492 & \textbf{0.511} & 0.456 
& \textbf{0.800} & 0.777 & 0.746 & 0.669 \\
Qwen3-VL-32B + prompt 
& 0.534 & 0.529 & \textbf{0.541} & 0.535 
& \textbf{0.875} & 0.839 & 0.835 & 0.639 \\
\Xhline{1.2pt}
\end{tabular}
\end{table}

\subsection{Additional qualitative examples} \label{sec:appendix-qualitative}

Figures~\ref{fig:qualitative-1-lasot},~\ref{fig:qualitative-2-pdm},~\ref{fig:qualitative-3-got10k}, and~\ref{fig:qualitative-4-vasttrack} show additional qualitative comparisons between IPLoc and IPLoc-ID using the Qwen3-VL-32B backbone.
Each 1-shot example consists of a reference image, a positive query image, and a negative query image.
The red BBOX denotes the reference annotation; in positive queries, blue and green BBOXes denote the ground truth and prediction, respectively; in negative queries, magenta denotes a false-positive detection.
These examples further illustrate that IPLoc-ID localizes the target object in positive queries while suppressing false positives in negative queries.

%
\def\customsize{\scriptsize}
\def\vspacesize{6pt}
\begin{figure*}[t]
\centering
\fbox{%
\begin{minipage}{\textwidth}
\vspace{3pt}
{\scriptsize\textbf{Instruction Prompt \#1}}\vspace{6pt}\\
{\customsize\ttfamily
SYSTEM: You are performing visual localization and identity verification.\par
Output exactly two lines.\par
Line 1: bbox=[x1,y1,x2,y2] in pixels for the LAST (target) image, inferred from the reference images/labels/bboxes.\par
Line 2: same\_object=YES or same\_object=NO indicating whether ALL boxes refer to the same object identity.\par
Do not output any other text.\par
\vspace{6pt}
}
\hrule\par
\vspace{\vspacesize}
{\scriptsize\textbf{Instruction Prompt \#2}}\vspace{6pt}\\
{\customsize\ttfamily
SYSTEM: Perform visual localization on the LAST (target) image using the reference images/labels/bboxes.\par
Return EXACTLY ONE LINE and NOTHING ELSE.\par
Output format must be: [x1, y1, x2, y2], YES\_or\_NO\par
- [x1, y1, x2, y2] are pixel coordinates for the target bbox.\par
- YES\_or\_NO is either YES or NO, indicating whether ALL boxes refer to the same object identity.\par
Do not include words like bbox=, same\_object=, in pixels, or any explanation.\par
\vspace{6pt}
}
\hrule\par
\vspace{\vspacesize}
{\scriptsize\textbf{Instruction Prompt \#3}}\vspace{6pt}\\
{\customsize\ttfamily
SYSTEM: Use the given reference images + labels + bboxes to localize the object in the last (target) image.\par
Then verify identity consistency across references and the target.\par
Finalize with exactly two lines (no extra text):\par
bbox=[x1,y1,x2,y2]\par
same\_object=YES/NO\par
\vspace{6pt}
}
\hrule\par
\vspace{\vspacesize}
{\scriptsize\textbf{Instruction Prompt \#4}}\vspace{6pt}\\
{\customsize\ttfamily
SYSTEM: First output a bbox for the target as [x1,y1,x2,y2]. Then answer the final identity question with only YES or NO.\par
Do not include explanations.\par
\vspace{6pt}
}
\end{minipage}%
}
\caption{\textbf{[Instruction prompts for general VLMs]} Concrete instruction prompts (\#1--\#4) used for joint box localization and identity verification in the prompt pretest.}
\label{fig:ablation:instruction-prompts}
\end{figure*}
%
%

\def \fw {160pt}
\def \pw {160pt}
\begin{figure}[htb]
\centering
\scriptsize
\begin{tabular}{cc}
\scriptsize \quad reference \qquad positive image \quad  negative image & 
\scriptsize \quad reference \qquad positive image \quad  negative image \\ 
\includegraphics[width=\fw]{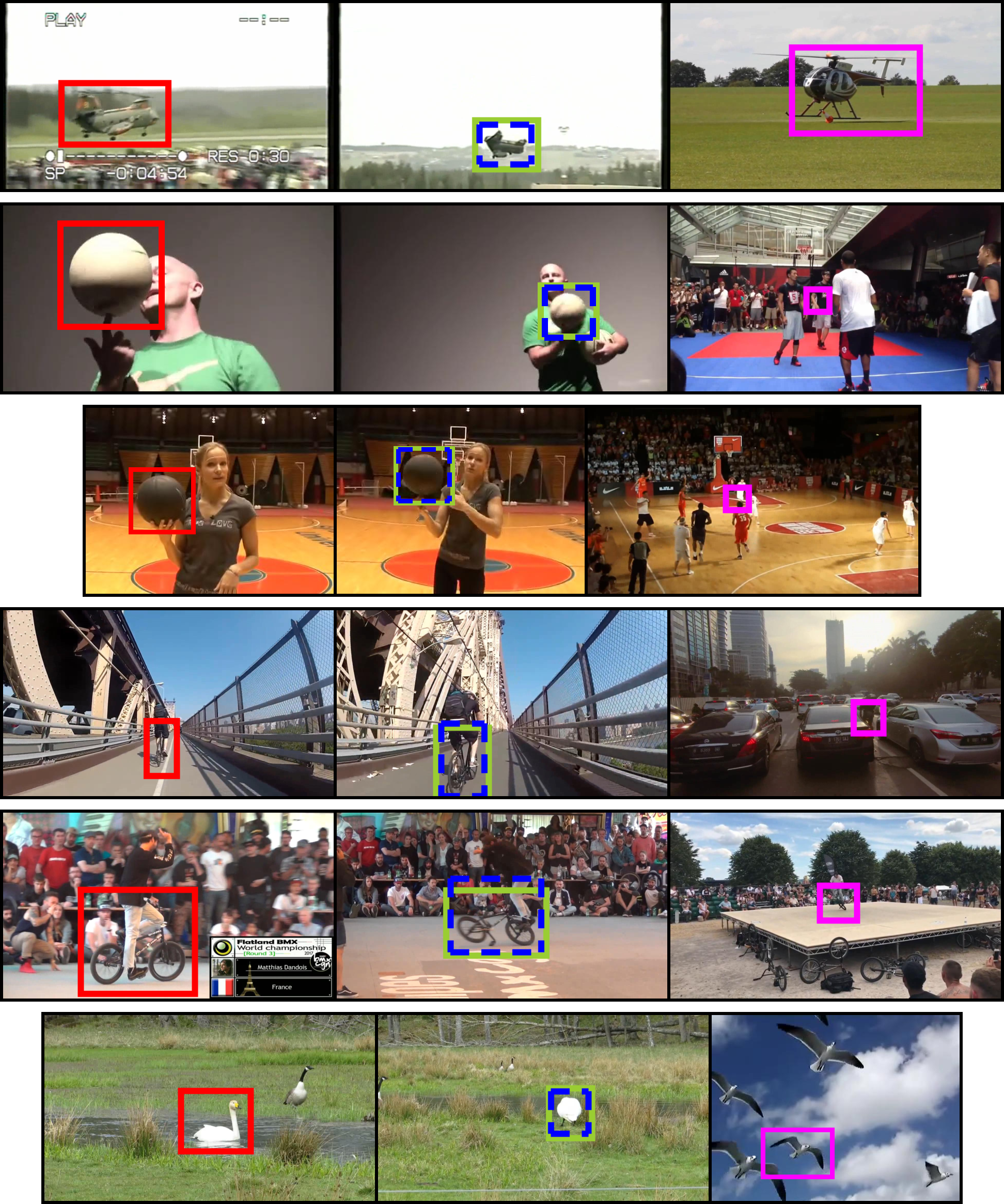} &
\includegraphics[width=\fw]{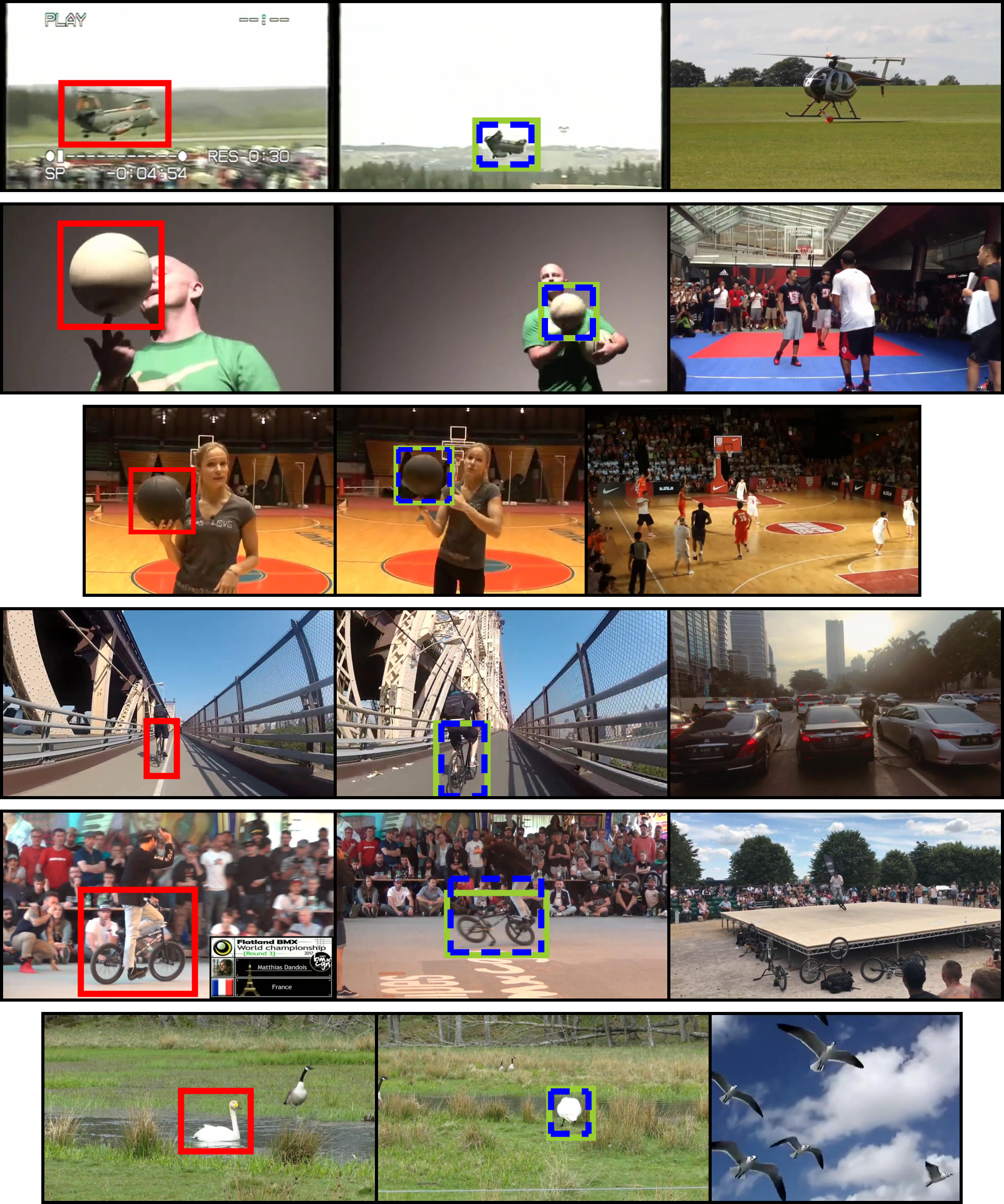} \\
(1) IPLoc & (2) IPLoc-ID \\
\end{tabular}
\vspace{-12pt}
\caption{\textbf{[Qualitative comparison on the LaSOT test set]}
Reference (red), true-positive (green) and false-positive (magenta) boxes using
IPLoc and IPLoc-ID.}
\label{fig:qualitative-1-lasot}
\end{figure}

\def \fw {160pt}
\def \pw {160pt}
\begin{figure}[htb]
\centering
\scriptsize
\begin{tabular}{cc}
\scriptsize reference \quad positive image \quad  negative image & 
\scriptsize reference \quad positive image \quad  negative image \\ 
\includegraphics[width=\fw]{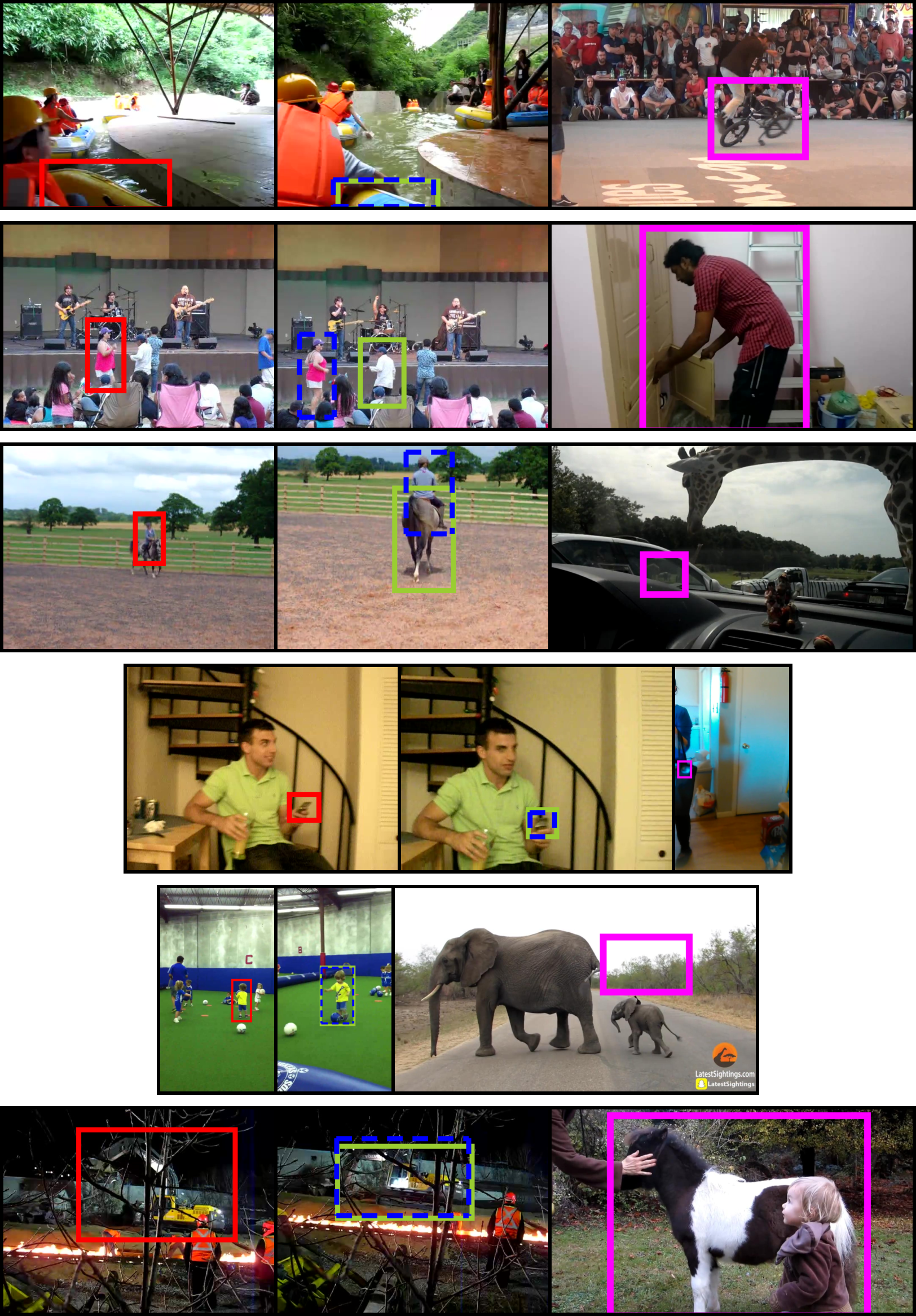} &
\includegraphics[width=\fw]{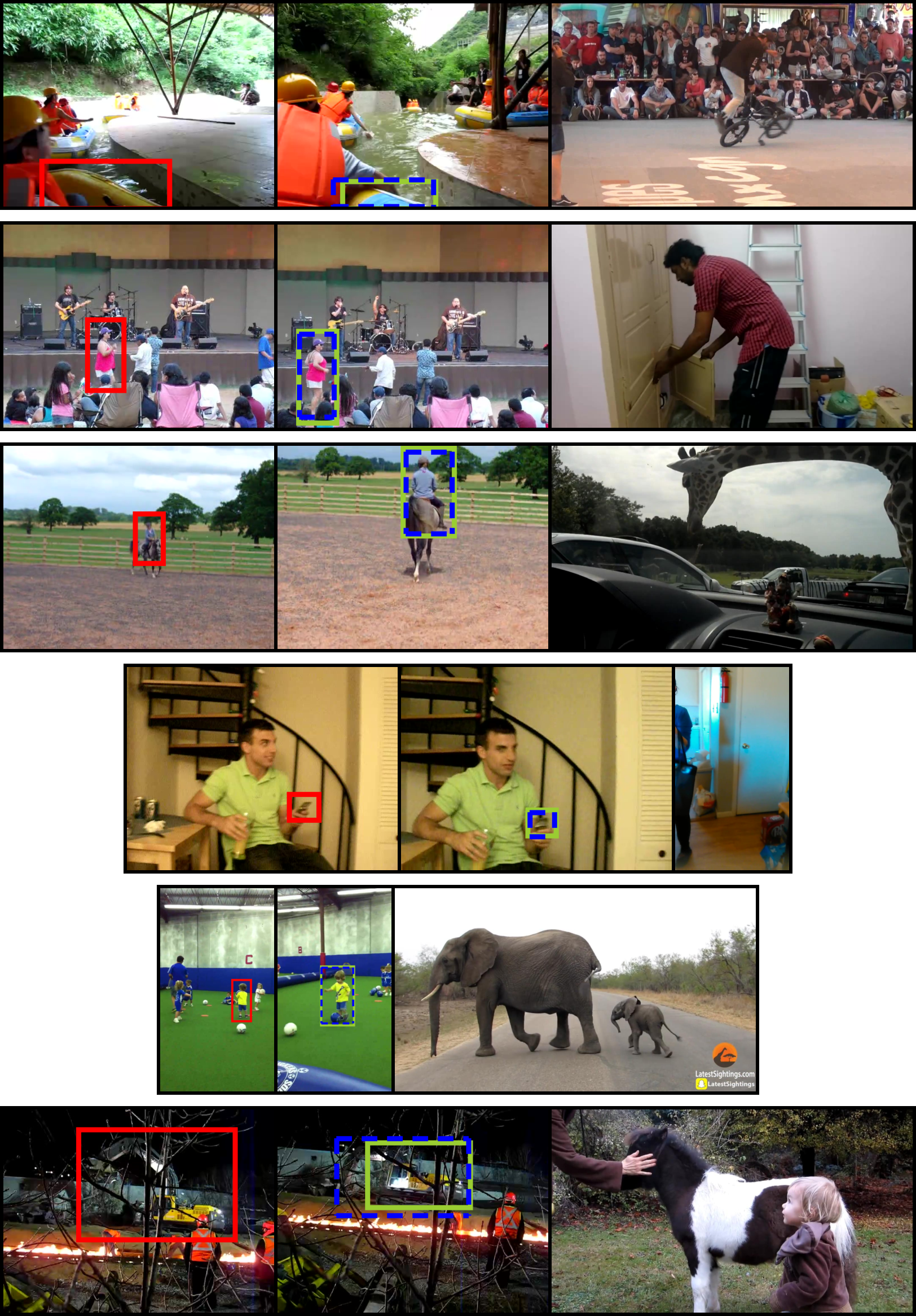} \\
(1) IPLoc & (2) IPLoc-ID \\
\end{tabular}
\vspace{-12pt}
\caption{\textbf{[Qualitative comparison on PDM]}
Reference (red), true-positive (green) and false-positive (magenta) boxes using
IPLoc and IPLoc-ID.}
\label{fig:qualitative-2-pdm}
\end{figure}

\def \fw {160pt}
\def \pw {160pt}
\begin{figure}[htb]
\centering
\scriptsize
\begin{tabular}{cc}
\scriptsize \quad reference \qquad positive image \quad  negative image & 
\scriptsize \quad reference \qquad positive image \quad  negative image \\ 
\includegraphics[width=\fw]{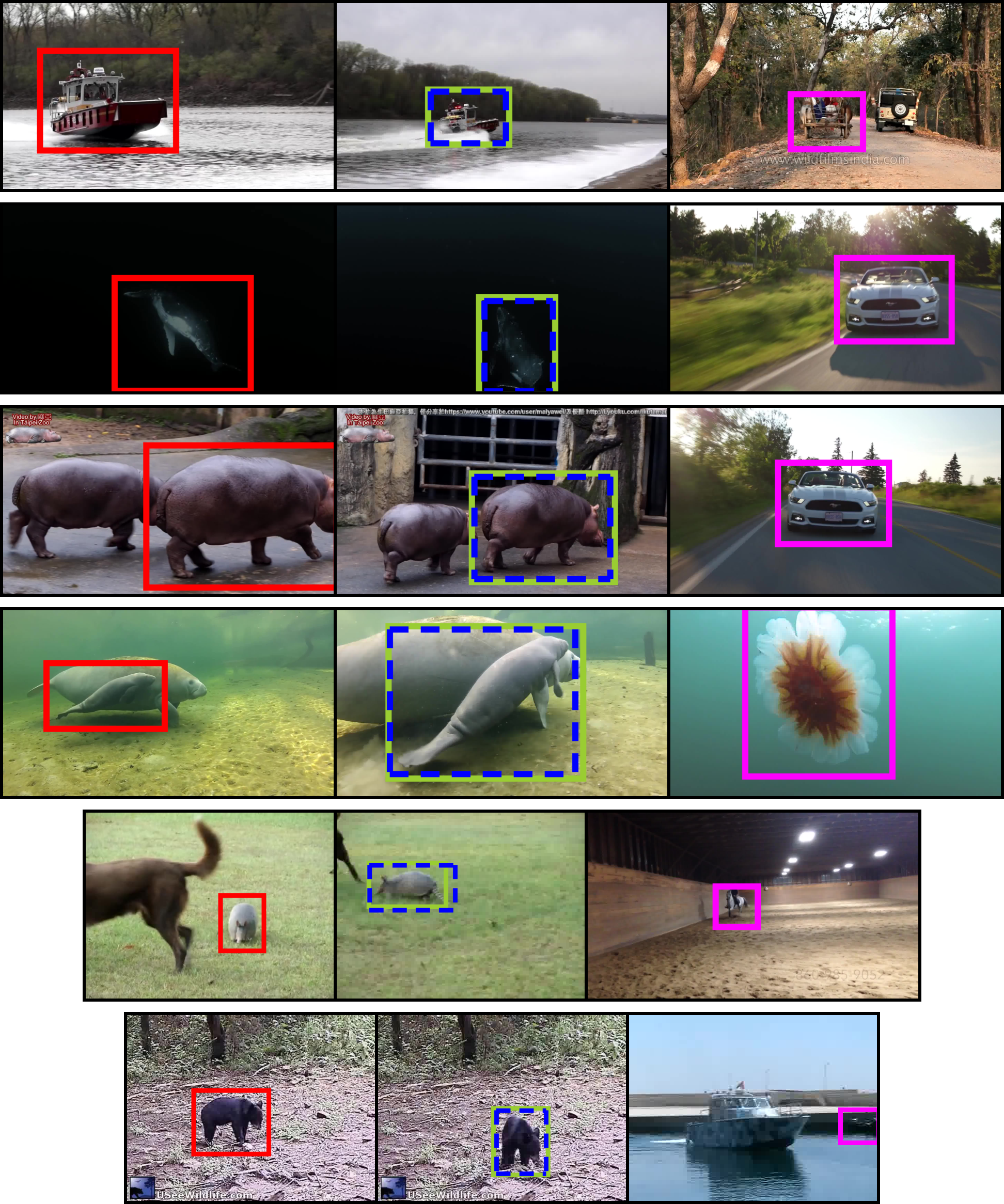} &
\includegraphics[width=\fw]{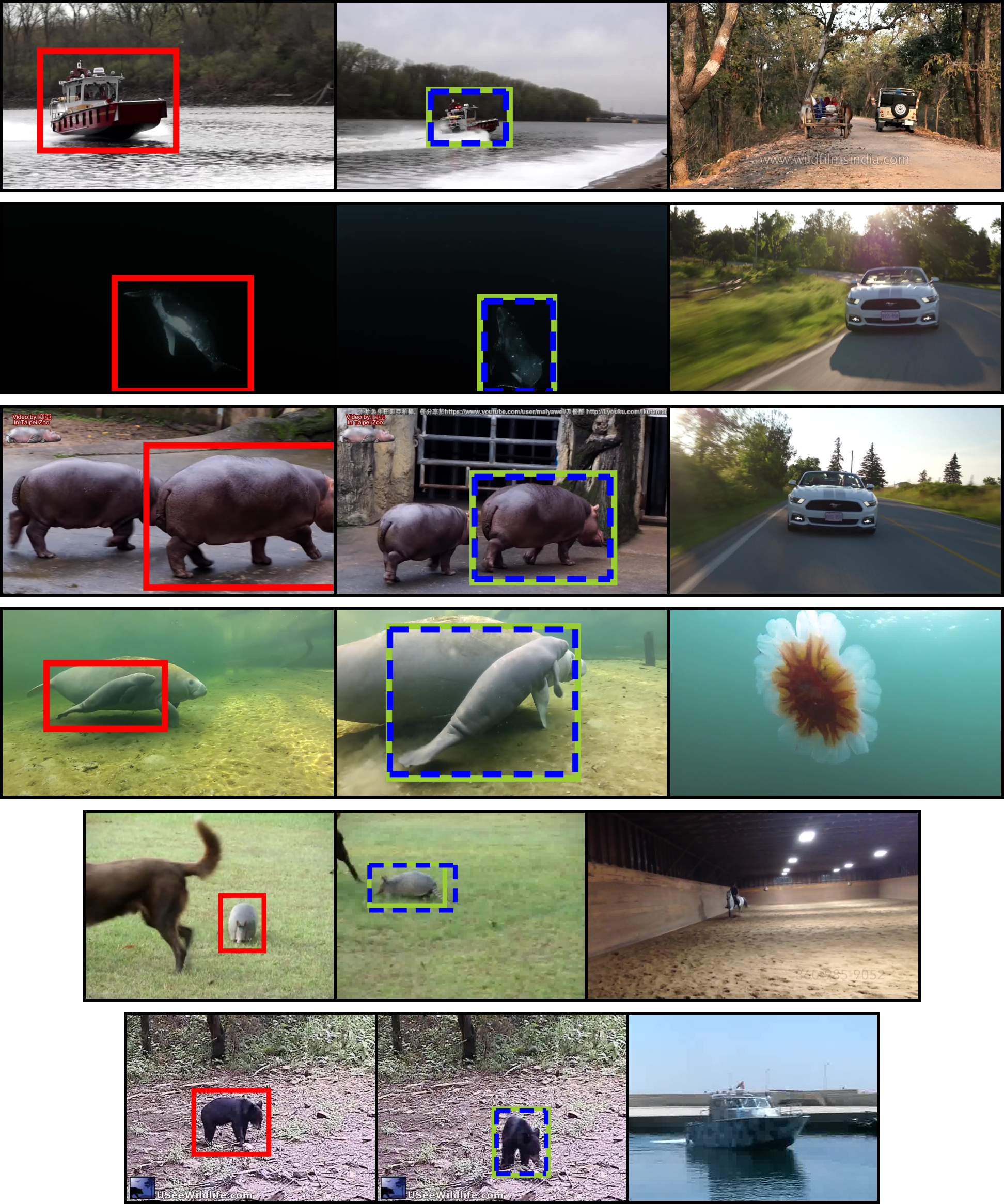} \\
(1) IPLoc & (2) IPLoc-ID \\
\end{tabular}
\vspace{-12pt}
\caption{\textbf{[Qualitative comparison on GOT-10K]}
Reference (red), true-positive (green) and false-positive (magenta) boxes using
IPLoc and IPLoc-ID.}
\label{fig:qualitative-3-got10k}
\end{figure}

\def \fw {160pt}
\def \pw {160pt}
\begin{figure}[htb]
\centering
\scriptsize
\begin{tabular}{cc}
\scriptsize \quad reference \qquad positive image \quad  negative image & 
\scriptsize \quad reference \qquad positive image \quad  negative image \\ 
\includegraphics[width=\fw]{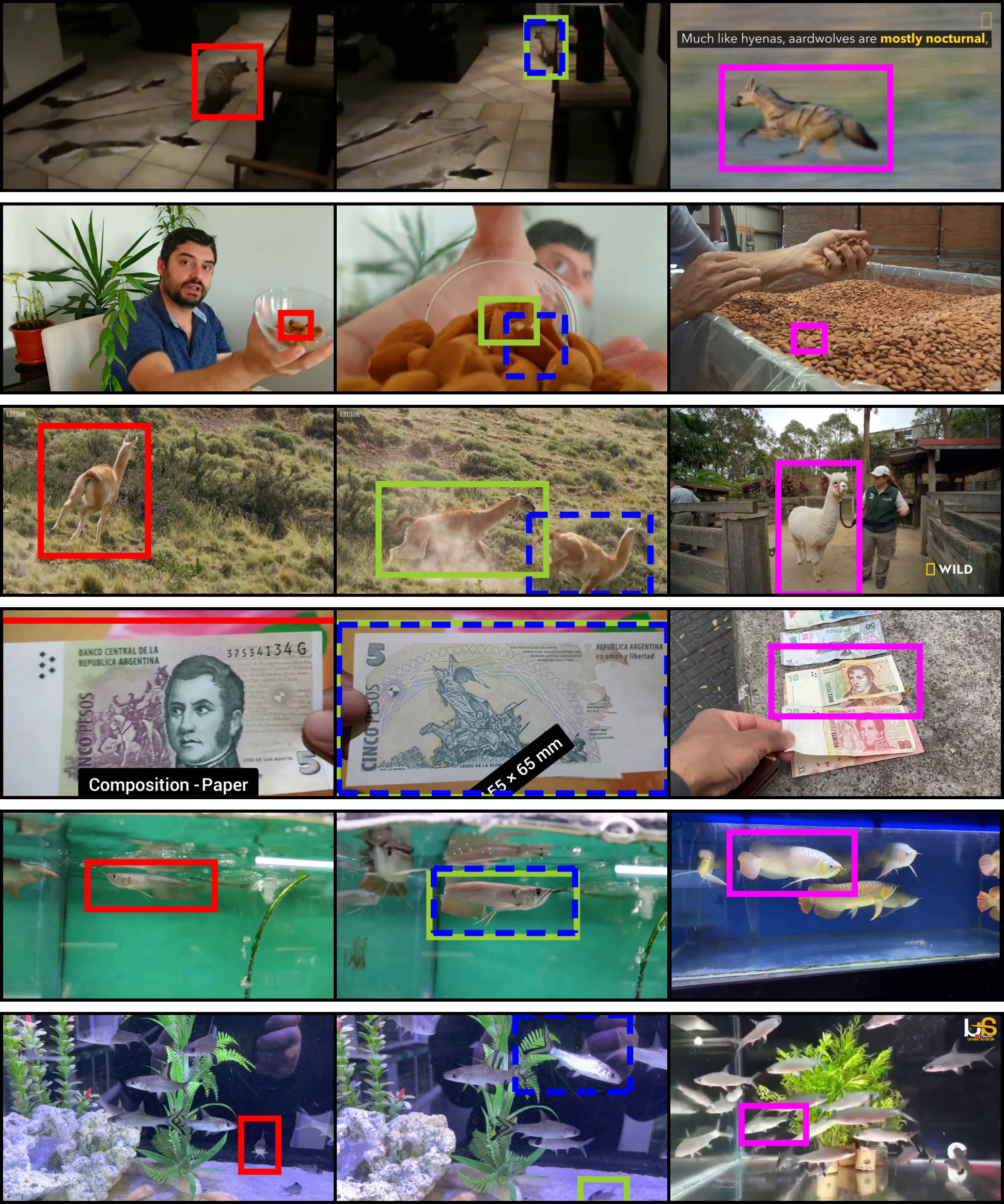} &
\includegraphics[width=\fw]{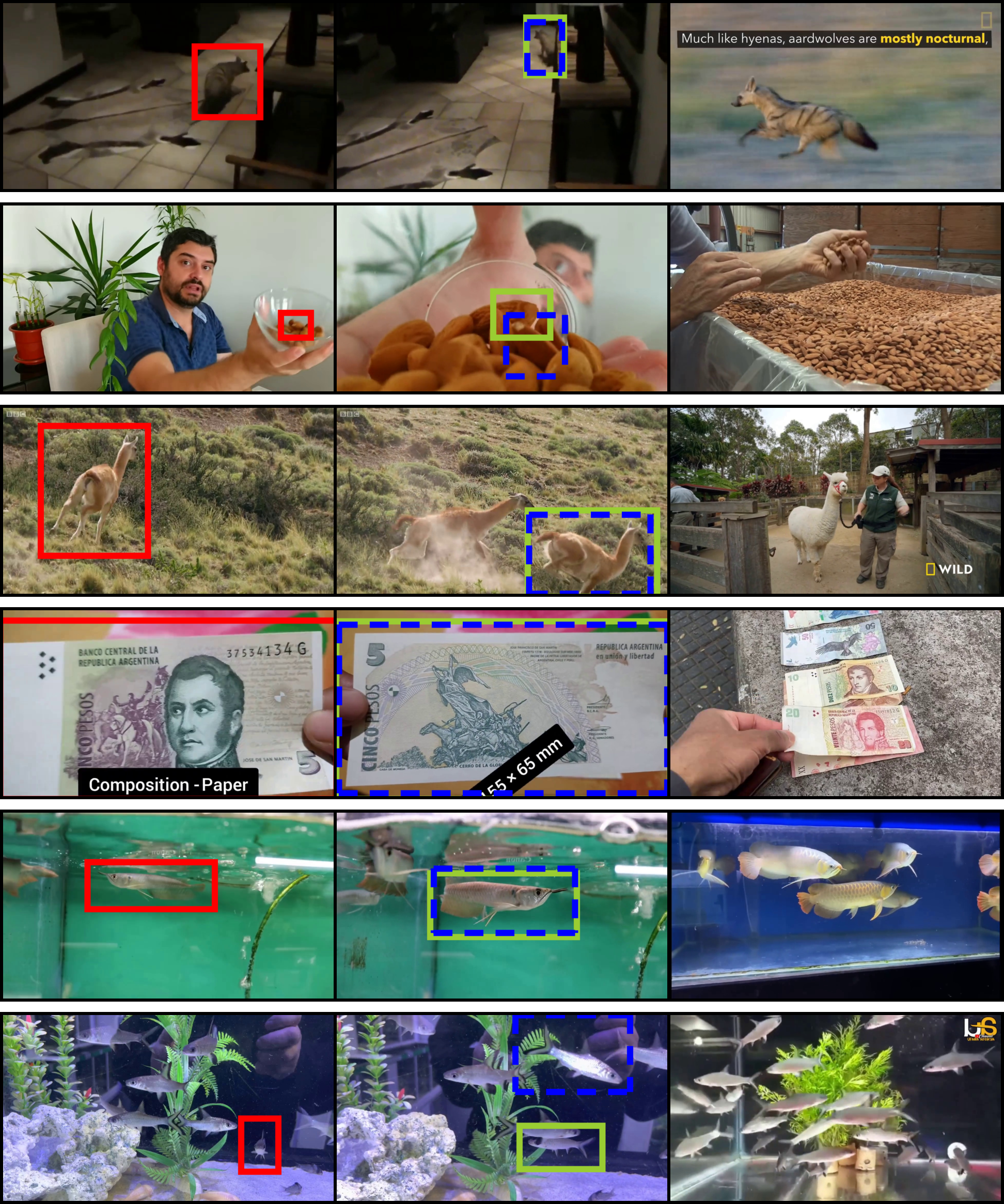} \\
(1) IPLoc & (2) IPLoc-ID \\
\end{tabular}
\vspace{-12pt}
\caption{\textbf{[Qualitative comparison on VastTrack]}
Reference (red), true-positive (green) and false-positive (magenta) boxes using
IPLoc and IPLoc-ID.}
\label{fig:qualitative-4-vasttrack}
\end{figure}

%
%
%
%
\end{document}